\PassOptionsToPackage{table}{xcolor}

\documentclass{article}

\usepackage{microtype}
\usepackage{graphicx}
\usepackage{subcaption}
\usepackage{booktabs} 
\usepackage{tabularx}

\usepackage[table]{xcolor}

\usepackage{colortbl}

\usepackage{hyperref}
\usepackage[preprint]{icml2026}

\usepackage{pifont}
\usepackage[most]{tcolorbox}

\definecolor{GrayRow}{gray}{0.92}

\definecolor{OpenSourceRow}{rgb}{1.0, 0.96, 0.90}

\definecolor{BlueRow}{rgb}{0.80, 0.90, 0.98}

\usepackage{amsmath}
\usepackage{amssymb}
\usepackage{mathtools}
\usepackage{amsthm}

\usepackage[capitalize,noabbrev]{cleveref}

\theoremstyle{plain}

\theoremstyle{definition}

\theoremstyle{remark}

\usepackage{enumitem}
\usepackage{multirow}
\usepackage{longtable}      
\usepackage{array}          
\usepackage{float}          

\usepackage[textsize=tiny]{todonotes}

\icmltitlerunning{
EnterpriseLab: A Full-Stack Platform for developing and deploying agents in Enterprises
}

\begin{document}

\twocolumn[
\icmltitle{EnterpriseLab: A Full-Stack Platform for developing and deploying agents in Enterprises}







\icmlsetsymbol{equal}{*}

\begin{icmlauthorlist}
  \icmlauthor{Ankush Agarwal}{equal,fujitsu}
  \icmlauthor{Harsh Vishwakarma}{equal,fujitsu}
  \icmlauthor{Suraj Nagaje}{equal,fujitsu}
  \icmlauthor{Chaitanya Devaguptapu}{fujitsu}
\end{icmlauthorlist}

\icmlaffiliation{fujitsu}{Fujitsu Research India}

\icmlcorrespondingauthor{Ankush Agarwal}{ankush.agarwal@fujitsu.com}
\icmlcorrespondingauthor{Harsh Vishwakarma}{harsh.vishwakarma@fujitsu.com}

  \icmlkeywords{Machine Learning, ICML}

  \vskip 0.3in
]

\newcommand{\name}{EnterpriseLab}



\printAffiliationsAndNotice{\icmlEqualContribution}  


\begin{abstract}
\vspace{-0.2cm}
Deploying AI agents in enterprise environments requires balancing capability with data sovereignty and cost constraints. While small language models offer privacy-preserving alternatives to frontier models, their specialization is hindered by fragmented development pipelines that separate tool integration, data generation, and training. We introduce \textbf{EnterpriseLab}, a full-stack platform that unifies these stages into a closed-loop framework. EnterpriseLab provides (1) a modular environment exposing enterprise applications via Model Context Protocol, enabling seamless integration of proprietary and open-source tools; (2) automated trajectory synthesis that programmatically generates training data from environment schemas; and (3) integrated training pipelines with continuous evaluation. We validate the platform through \textbf{EnterpriseArena}, an instantiation with 15 applications and 140+ tools across IT, HR, sales, and engineering domains. Our results demonstrate that 8B-parameter models trained within EnterpriseLab match GPT-4o's performance on complex enterprise workflows while reducing inference costs by 8-10×, and remain robust across diverse enterprise benchmarks, including EnterpriseBench (+10\%) and CRMArena (+10\%). EnterpriseLab provides enterprises a practical path to deploying capable, privacy-preserving agents without compromising operational capability.
The blog containing demo videos, code, and data is available at \href{https://ast-fri.github.io/EnterpriseLab/}{EnterpriseLab}.

\end{abstract}
\vspace{-0.2cm}
\section{Introduction}
\vspace{-0.2cm}
\begin{figure*}[htbp]
    \centering
    \includegraphics[width=0.9\textwidth]{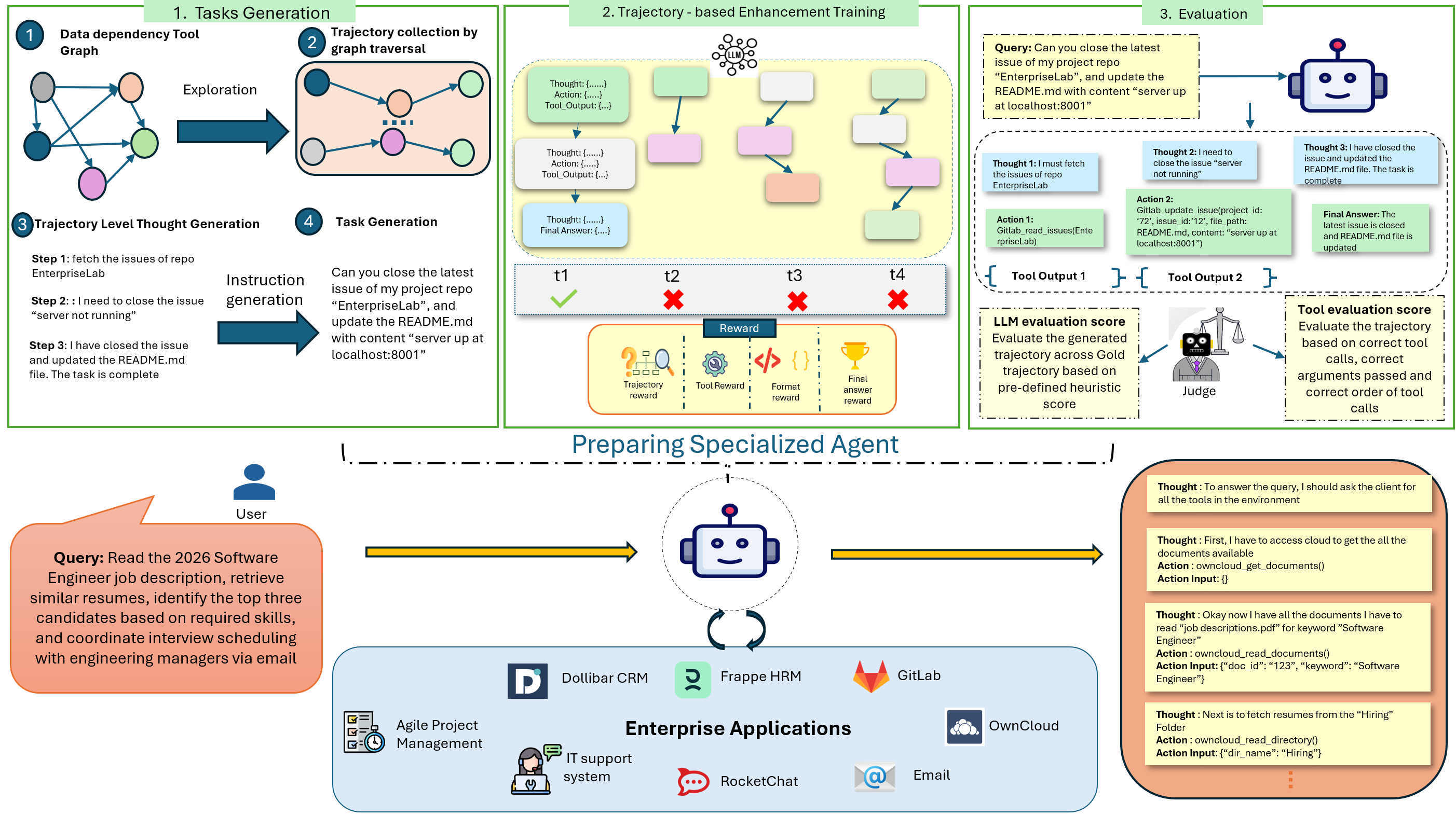}
    \caption[EnterpriseLab]{\textbf{Full Stack Platform for Developing Enterprise Agents:} The \name{} modules (1, 2, and 3) collaborate to create specialized agents, which are then deployed to the agentic environment for end-user interaction.}
    \label{fig:ELab}
    \vspace{-0.4cm}
\end{figure*}


LLM based AI agents have become a cornerstone in improving productivity at enterprises~\cite{glean2025platform}. Enterprises require intelligent automation across complex, cross-departmental workflows spanning HR, IT, sales, and engineering operations. While frontier language models such as GPT-4o~\cite{hurst2024gpt}, Claude~\cite{anthropiclaude}, and Gemini~\cite{gemini3} demonstrate strong reasoning and tool-use capabilities~\cite{li2024humaneval, zhaocommit0, pantraining, yangswe}, their deployment in enterprise settings faces critical constraints. Data sovereignty regulations, high inference costs (\$3-\$15 per million tokens), and API latency hinder their adoption. Small Language Models (SLMs) in the 8B-32B parameter range offer a promising alternative: they enable on-premises deployment, reduce inference costs by an order of magnitude, and provide fine-grained control over model behavior~\cite{belcak2025small}. However, while model architecture plays a role, effective specialization to enterprise workflows is primarily constrained by \textit{infrastructure}, namely the absence of integrated systems that can transform internal tools and business logic into high-quality training data.

The infrastructure gap manifests in fragmented development pipelines that separate tool integration, data collection, and model training into disconnected stages. Existing agentic benchmarks such as CRMArena~\cite{huang2025crmarena}, EnterpriseBench~\cite{vishwakarma2025can}, and WebArena~\cite{zhouwebarena} provide valuable evaluation suites, but they are designed to \textit{measure} agent performance, not to \textit{build} agents-they use static task sets, do not connect to live enterprise tool stacks, and do not generate training data. Separately, recent data synthesis work including ToolACE~\cite{liutoolace} and Graph2Eval~\cite{chen2025graph2eval} produces training trajectories but operates independently of execution environments, preventing environment feedback from informing synthesis and eliminating online learning capabilities. Consequently, organizations attempting to deploy specialized agents face substantial barriers: custom integration of heterogeneous tools, manual annotation of interaction trajectories, and the absence of unified infrastructure for iterative development~\cite{warrier2023managing, marro2025llm, bodensohn2025unveiling}.

We introduce \textbf{EnterpriseLab}, a full-stack platform that unifies tool integration, data synthesis, model training, and evaluation into a closed-loop framework for enabling development of AI agents in enterprise contexts. EnterpriseLab is architected around three tightly integrated components: (1) a \textit{modular tool environment} that exposes enterprise applications via Model Context Protocol (MCP)\footnote{\url{https://modelcontextprotocol.io/specification/2025-03-26}}, enabling plug-and-play integration of proprietary and open-source tools; (2) an \textit{automated trajectory synthesis pipeline} that programmatically generates executable training data from environment schemas via constraint-aware tool graph traversal; and (3) an \textit{integrated training infrastructure} supporting supervised fine-tuning, preference optimization, and online reinforcement learning with continuous evaluation. This closed-loop design ensures that model training receives direct feedback from tool execution, enabling rapid iteration as enterprise workflows evolve.

To validate EnterpriseLab's design, we instantiate it as \textbf{EnterpriseArena},  a comprehensive environment with 15 containerized applications exposing 140+ tools across IT, HR, sales, engineering, and communication domains. Applications  are initialized with realistic synthetic data, creating stateful dependencies where actions propagate across systems (e.g., HR employee records trigger CRM assignments). EnterpriseArena includes 500 expert-curated evaluation tasks requiring 3-12 tool invocations across 2-5 servers. While EnterpriseArena uses  open-source tools for reproducibility, EnterpriseLab supports arbitrary tool integration, enabling enterprises to substitute proprietary systems.

Consider this cross-functional task: \textit{``Read the 2026 Software Engineer job description, fetch relevant resumes, identify the top three candidates based on required skills, and coordinate interview scheduling with engineering managers via email.''} This workflow requires orchestrating HR systems, document storage, and communication platforms with stateful reasoning where skills guide candidate ranking. EnterpriseLab handles this naturally: the modular environment provides MCP servers for each application with realistic data; the synthesis pipeline constructs tool dependency graphs and  generates similar tasks by traversing valid execution paths; and training provides  direct feedback from tool execution. Evaluation occurs within the same environment, ensuring metrics reflect genuine operational capability.

Our empirical evaluation demonstrates that with unified infrastructure, small models can match the performance of proprietary models in enterprise settings. A Qwen3-8B model~\cite{yang2025qwen3} trained on 500 synthesized trajectories within EnterpriseLab achieves a 30\% improvement in execution accuracy over its base version and matches the performance of GPT-4o on EnterpriseArena while reducing inference costs by 8-10x (Tables~\ref{tab:exec_results} \& ~\ref{tab:cost_appendix}).  Cross-environment validation shows models trained via EnterpriseLab outperform GPT-4o by 10\% on both EnterpriseBench~\cite{vishwakarma2025can} and CRMArena~\cite{huang2025crmarena}, with competitive performance on $\tau$-Bench~\cite{yao2025tau}, demonstrating  strong adaptability across diverse task environments. Our training pipeline scales efficiently: supervised fine-tuning completes within 2 hours, while online RL training requires 24-30 hours on 4xH200 GPUs, yielding production-ready models from raw tool schemas in under two days. 
\\
\noindent These results show that with appropriate training infrastructure, compact models can achieve frontier-level performance on enterprise tasks, enabling organizations to deploy cost-effective, privacy-preserving agents. Our contributions are summarized as follows:
\vspace{-0.2cm}
\noindent
\begin{itemize}[noitemsep, wide=0pt]
    \item \textbf{EnterpriseLab}, a full-stack platform that integrates tool connectivity, trajectory synthesis, model training, and evaluation into a unified closed-loop framework, enabling enterprises to transform proprietary tools and workflows into deployable agents without external dependencies.
    
    \item \textbf{EnterpriseArena}, a comprehensive instantiation of EnterpriseLab comprising 15 containerized applications exposing 140+ tools across IT, HR, sales, engineering, and communication domains, with 500 expert-curated multi-step evaluation tasks, demonstrating the platform's capability to simulate realistic cross-departmental enterprise workflows.
    
    \item Automated trajectory synthesis via constraint-aware tool graph traversal, which programmatically generates executable training data from environment schemas, eliminating manual annotation while ensuring data-flow validity and task diversity.
    
   \item Empirical validation demonstrating that 8B-parameter models trained within EnterpriseLab match GPT-4o's performance on complex enterprise tasks while reducing inference costs by 8--10$\times$, and achieve consistent performance gains on EnterpriseBench (+10\%) and CRMArena (+10\%), providing enterprises a practical alternative to proprietary API dependence.

\end{itemize}

\vspace{-0.2cm}
\section{The EnterpriseLab Platform}
\vspace{-0.2cm}
We present \name{}, a unified platform for training enterprise AI agents through closed-loop integration of tool environments, data synthesis, and training infrastructure. The platform comprises three components: a modular environment architecture (Section~\ref{sec:modular_env}), an automated task synthesis pipeline (Section~\ref{sec:scapia-data}), and an integrated training infrastructure with trajectory-level optimization (Section~\ref{sec:scapia-training}). Figure~\ref{fig:ELab} illustrates the system architecture.

\vspace{-0.2cm}
\subsection{Modular Tool Environment Architecture}
\label{sec:modular_env}
\vspace{-0.2cm}
We implement the environment layer as a client-server system where agents interact with dynamically connected enterprise applications. The architecture comprises three components: (i) a dynamic tool registry that queries active servers at runtime to discover available tools and constructs unified action schemas by normalizing parameter names, types, and descriptions into a consistent JSON format—for example, mapping \texttt{repository} (GitHub) and \texttt{project} (Jira) to a standard \texttt{workspace\_id} field while preserving tool-specific namespaces (e.g., \texttt{github.id}, \texttt{jira.id}) to resolve semantic conflicts; (ii) stateful execution containers that map each training episode to a dedicated Docker instance with persistent storage, maintaining file systems, database states, and authentication tokens across multi-turn trajectories; and (iii) an observation normalizer that captures heterogeneous tool outputs (structured API responses, command-line streams, error logs) and transforms them into a token-budget JSON format with importance-based truncation prioritizing error messages and return values over verbose logs. While our implementation uses MCP-compliant servers for standardized tool discovery and invocation, the architecture supports non-MCP tools through adapter wrappers. Enterprise applications can be added or removed by launching or terminating their corresponding server processes, requiring no modifications to the agent code or training pipeline. Section~\ref{sec:entarena} describes EnterpriseArena, a benchmark instantiation comprising 15 enterprise applications and 140+ tools.

\vspace{-0.2cm}
\subsection{Task Synthesis Pipeline}
\label{sec:scapia-data}
\vspace{-0.2cm}
To mitigate the reliance on manually annotated enterprise workflows, we propose an automated pipeline that synthesizes high-quality, executable tasks from the environment itself. Formally, we define an environment as a tuple $\mathcal{E} = (\mathcal{T}, \mathcal{S})$, where $\mathcal{T}$ is the set of available tools and $\mathcal{S}$ is the state space that evolves through tool executions.
 Our objective is to generate a dataset $\mathcal{D} = \{(x_i, t_i)\}_{i=1}^N$, where $x_i$ is a natural language intent and $t_i = [t_1, \dots, t_L]$ is a valid sequence of tool invocations such that each step $t_j \in \mathcal{T}$ is executable and meaningful. To achieve this, we organize our generation process into four phases:

\textbf{Phase 1: Tool Graph Construction.} We first aggregate all tools exposed by the environment into a unified registry $\mathcal{T}$, including their argument schemas (names, types, required/optional fields, defaults) and, when available, return schemas. For static environments, this registry is loaded from configuration artifacts (e.g., JSON/YAML), whereas for dynamic environments we query MCP servers via their tool-listing interfaces. We normalize these heterogeneous definitions into a consistent internal representation and cache the registry for reuse within an episode. We then model the tool space as a directed dependency graph $G_h = (\mathcal{T}, E)$, where a directed edge $(t_i, t_j) \in E$ is added if a return field of $t_i$ is type-and-name compatible with a required input argument of $t_j$. This graph encodes data-flow feasibility, ensuring that any path in $G_h$ corresponds to a sequence in which required inputs can be satisfied by prior outputs.

\textbf{Phase 2: Constraint-Aware Trajectory Sampling.}
Given the constructed tool graph $G_h$, we perform depth-first traversal up to a maximum depth $L$  or breaking at first visited node, starting from valid entry nodes which includes: (i) \textsc{CREATE} tools that instantiate new entities, (ii) \textsc{LIST/SEARCH} tools with no mandatory input dependencies, and (iii) tools whose required inputs can be satisfied by environment-provided seed data. This selection is to ensure that the input arguments provided during exploration are not hallucinated and obeys environment schema.
During traversal, we maintain two memory buffers: a local trajectory memory $\mathcal{M}_{\text{local}}$ that stores outputs produced along the current path, and a global memory $\mathcal{M}_{\text{global}}$ that aggregates entities across previously explored trajectories. The input arguments are either generated by LLM (for CREATE tools) or fetched from these memory buffers so that on encountering a \textsc{READ/UPDATE/DELETE} tool, even if a required argument is not present in parent tool's output, can be fetched from the memory with priority given to $\mathcal{M}_{\text{local}}$ to preserve within-trajectory consistency.
Trajectory expansion terminates upon reaching depth $L$, encountering a leaf node or a visited node, or when no successor tool has satisfiable inputs. For each starting node, we collect up to $K$ valid trajectories, yielding a diverse set of executable tool sequences used for downstream task generation.

\textbf{Phase 3: Hierarchical Task Synthesis.}
For each trajectory, we enumerate all contiguous subsequences $\ell \in [2, L]$. We first prompt an LLM to generate low level thought for consecutive pair of nodes in subsequence, conditioned on tool names, argument schemas, and trajectory memory outputs, such that the corresponding tool sequence constitutes a valid solution. We then prompt again to synthesize high-level user intents that describe the full trajectory by composing these low-level semantics into a single coherent request (e.g., `create\_repo $\rightarrow$ add\_file` becomes ``Set up a new project''). 
This two-stage strategy, adapted from \citet{sun2025genesis}, produces tasks spanning fine-grained sub-tasks to multi-step objectives. Low-level thought and high-level thought prompts are present in Appendix \ref{app:thought_prompt} and \ref{app:high_prompt} respectively.

\textbf{Phase 4: Validation and Filtering.}
The raw task set undergoes a three-stage validation pipeline: (i) de-duplication via exact matching and fuzzy string similarity (threshold $\geq$ 0.9) to remove duplicate and near-duplicate task descriptions; (ii) diversity-based filtering using Maximal Marginal Relevance (MMR)~\cite{goldstein1998summarization} with cosine similarity on sentence embeddings to iteratively select tasks that maximize semantic distance from already-selected tasks while maintaining coverage of distinct trajectory patterns; and (iii) grounding validation by executing each task's reference trajectory in the environment and verifying that all tool calls return successful execution status and outputs conforming to expected schemas. Tasks that fail execution are discarded, yielding a high-quality, diverse, and executable task corpus.

\vspace{-0.2cm}
\subsection{Integrated Training Infrastructure}
\label{sec:scapia-training}
\vspace{-0.2cm}
\name{} transforms synthesized task corpus into trained agents capable of multi-tool coordination through supervised fine-tuning, preference alignment, and trajectories optimization. We first describe the agent scaffolding and execution infrastructure that standardizes tool interaction across training paradigms. Next, we detail offline training methods that optimize policies from static trajectories. Finally, we present Agentic GRPO, an online reinforcement learning method that directly incorporates environment interaction into policy optimization.

\textbf{Agent Scaffolding and Execution.}
\label{sec:scaffolds}
\name{} supports multiple agent scaffolding strategies for executing tool-calling workflows, with prompting techniques such as ReAct~\cite{yao2022react} applicable to both open-weight models (e.g., Qwen~\cite{yang2025qwen3}, Llama~\cite{touvron2023llama}) and proprietary models (e.g., GPT~\cite{gpt5_o3}, Claude~\cite{claude_sonnet}), where the agent alternates between reasoning and tool execution steps using example trajectories sourced and validated through tasks synthesis pipeline. For proprietary models, \name{} additionally interfaces via native API-based function calling with structured tool schemas. All agent executions are logged and cached, providing trajectory data for both offline training and online policy improvement.

\textbf{Offline Training Methods.}
\label{sec:offline-methods}
Given a static task corpus and cached trajectories, \name{} supports two offline training regimes. SFT trains agents via cross-entropy loss on expert trajectories, with support for full fine-tuning and LoRA~\cite{hu2022lora} for parameter-efficient adaptation. DPO~\cite{rafailov2023direct} enables preference-based alignment by constructing (chosen, rejected) pairs from successful and failed trajectory rollouts, optimizing the policy to favor high-quality tool sequences without explicit reward modeling. These methods provide strong baselines but do not adapt to environment feedback during training.

\textbf{Agentic GRPO.}
\label{sec:agentic-grpo}
We apply Group Relative Policy Optimization (GRPO)~\cite{shao2024deepseekmath} in an agentic RL setting~\cite{singh2025agentic}, where trajectories are generated via ReAct-style rollouts (Algorithm~\ref{alg:agentic_grpo}). The prompt template for generating rollouts for Agentic GRPO is provided in Appendix~\ref{app:agrpo_prompt}. Following prior agentic RL and ARTIST approaches, tokens originating from tool outputs are masked during loss computation to prevent spurious gradients from deterministic environment responses and to focus optimization on agent reasoning and decision-making.
\\
\noindent For each query $\mathbf{q}$, we sample $G$ trajectories $\{\tau_i\}_{i=1}^G$ in environment $\mathcal{E}$ and obtain scalar trajectory rewards $r(\tau_i) \in [0,1]$ via environment execution, reflecting task completion, tool selection correctness, tool execution success, and answer validity (full details in Appendix \ref{app:reward}). Group-relative advantages are computed as $\hat{A}_i = \frac{r(\tau_i) - \bar{r}_G}{\sigma_G + \epsilon}$, where $\bar{r}_G$ and $\sigma_G$ denote the group mean and standard deviation. All tokens within a trajectory $\tau_i$ share the same advantage, enabling stable credit assignment in multi-turn tool-calling episodes, where intermediate steps cannot be evaluated independently.

\vspace{-0.2cm}

\section{EnterpriseArena: Benchmark Instantiation}
\label{sec:entarena}
\vspace{-0.2cm}
EnterpriseArena is a concrete instantiation of the modular environment architecture (Section~\ref{sec:modular_env}), serving as a comprehensive benchmark for evaluating agentic AI systems on realistic enterprise workflows. The benchmark comprises 15 specialized MCP servers emulating production applications and 500 expert-curated tasks spanning cross-functional enterprise scenarios.

\textbf{MCP Server Ecosystem.} EnterpriseArena comprises 15 specialized MCP servers that emulate production enterprise applications, collectively exposing 140+ tools spanning communication, development, operations, human resources, data storage, and business domains. Table~\ref{tab:mcp_servers} provides an ecosystem overview, while Appendix~\ref{app:mcp_desc} details comprehensive tool specifications. These servers are populated with simulated enterprise data adapted from EnterpriseBench~\cite{vishwakarma2025can}, subsequently verified and extended by nine domain experts (Table~\ref{tab:profession_data}).

Unlike static benchmarks, EnterpriseArena relies on a unified stateful backend where data changes propagate automatically across server boundaries. For instance, creating an HR employee record updates the central registry, which programmatically enables subsequent CRM assignments and notification dispatches without external intervention. The environment enforces realistic enterprise constraints through API-level validation and strict workflow dependencies.

\textbf{Task Design and Complexity.} EnterpriseArena contains 500 expert-curated tasks requiring multi-step planning, cross-application orchestration, and context-aware decision making. 
Tasks span eight enterprise domains with the distribution shown in Table~\ref{tab:domain_dist}. Each task is classified into one of five workflow categories: (1) CRUD operations, (2) Search \& Orchestration, (3) Multi-entity Workflow, (4) Version Control, and (5) Cross-functional Integration. These tasks were developed through structured Google Sheets based reviews with industry experts (see Figure~\ref{fig:cluster} in the Appendix).

\textbf{Task Examples.} For instance, the task \textit{``Create a project named `Nexus', add frontend, backend, and database modules, and create issues in each module"} requires sequential Plane MCP operations followed by team notification via RocketChat. A more complex example spanning multiple domains: \textit{``Read job description from Documents, fetch shortlisted resumes, identify top three candidates, and coordinate interview scheduling with aarav.mittal and aakas.bhalla"} orchestrates OwnCloud, Frappe HR, and Mail System for document retrieval, candidate ranking, and communication. Table~\ref{tab:complex_task_examples} presents additional representative examples with specific tool sequences.
Each task includes gold-standard trajectories hand-crafted by domain experts, comprising: (i) optimal tool call sequences with exact parameters, (ii) expected intermediate environment states, (iii) validation checkpoints for workflow milestones, and (iv) acceptable alternative paths where multiple valid solutions exist.

\vspace{-0.2cm}

\section{Experimental Setup}
\vspace{-0.2cm}
We evaluate \name{} by training a specialized agent in EnterpriseArena and testing its adaptability across three additional multi-turn tool-calling benchmarks. Below we describe the benchmark environments, baseline methods and their agentic configurations, evaluation metrics, and implementation details.

\vspace{-0.2cm}
\subsection{Benchmarks}
\vspace{-0.2cm}
We evaluate \name{} on EnterpriseArena, the benchmark introduced in this work (Section~\ref{sec:entarena}), which spans HR, Software Engineering, and IT operations domains. To assess adaptability beyond our platform, we also evaluate on three additional multi-turn tool-calling benchmarks.
EnterpriseBench~\cite{vishwakarma2025can} contains 500 instances across five enterprise domains such as Business and Sales. Tasks involve document analysis, data extraction, and domain-specific reasoning.  
CRMArena~\cite{huang2025crmarena} comprises 1,170 customer service queries spanning three personas: Service Manager, Service Agent, and Service Analyst. Agents must query and manipulate Salesforce data to resolve customer issues.  
$\tau$-Bench~\cite{yao2025tau} evaluates agents on retail and airline customer service scenarios with 165 test instances. Tasks involve multi-turn dialogues where agents interact with simulated users through tool calls.
We generate task-specific training data for each benchmark as explained in Section \ref{sec:scapia-data}. Table~\ref{tab:experiment_data} in Appendix summarizes the generated training tasks and the test set coverage for each environment.

\vspace{-0.2cm}
\subsection{Methods and Variants}
\label{sec:methods}
\vspace{-0.2cm}
For all benchmarks, we adopt the ReAct framework~\cite{yao2022react} as the agentic execution pipeline, in which agents interleave reasoning traces with tool actions. For EnterpriseBench, $\tau$-Bench, and CRMArena, we use the ReAct configurations provided in the original benchmarks. For EnterpriseArena, we implement a standard ReAct pipeline. In all cases, the corresponding ReAct pipelines are used to generate agent trajectories for evaluation.
\\
\noindent We evaluate Qwen3-8B~\cite{yang2025qwen3} in four configurations:
(1) the base pretrained model,  
(2) supervised fine-tuning with LoRA adapters~\cite{hu2022lora} on synthetic trajectories generated by \name{},  
(3) Direct Preference Optimization ~\cite{rafailov2023direct} trained on preference pairs constructed by sampling trajectories from both the base and SFT models and ranking them by task success, and  
(4) Agentic GRPO (Section~\ref{sec:agentic-grpo}) trained using Algorithm~\ref{alg:agentic_grpo}.
\\
\noindent We compare these models against two state-of-the-art tool-calling models: ToolACE~\cite{liutoolace}, a Llama-3.1-8B model trained on synthetic data generated from an API pool of 26,507 diverse APIs across 30 domains, and xLAM-2-70B~\cite{prabhakar2025apigen}, a Llama-3.1-70B model trained on 60k function-calling instances from APIGen and additional multi-turn data from APIGen-MT. Additionally, we evaluate three proprietary models with few-shot prompting: GPT-4o\footnote{\label{gpt4o}\url{https://platform.openai.com/docs/models\#gpt-4o}}, Claude-3.5-Sonnet\footnote{\label{claude}\url{https://aws.amazon.com/bedrock/claude/}}, and Gemini-2.5-Pro\footnote{\label{gemini}\url{https://ai.google.dev/gemini-api}}.

\vspace{-0.2cm}
\subsection{Evaluation Metrics}
\label{sec:metrics}
\vspace{-0.2cm}
We evaluate using MCPEval~\cite{liu2025mcpeval}, a two-phase framework combining tool execution accuracy and expert judgment. Phase-1 matches predicted tool calls against ground-truth trajectories by tool name, parameters, and execution order using \textit{strict matching} (exact agreement) or \textit{flexible matching} (similarity thresholds: parameters $\geq0.6$, execution order $\geq0.5$), reporting \textit{flexible matching} unless otherwise stated. Phase-2 uses GPT-4o to score trajectory quality including planning, execution flow, tool selection and usage, adaptability, efficiency, and context awareness and task completion quality including coverage, accuracy, completeness, and usefulness each in $[0,1]$. Phase 2 applies across all benchmarks while Phase 1 applies only to EnterpriseBench and EnterpriseArena, which provide gold tool sequences. For entity-creation tasks in these benchmarks, we execute a \texttt{read()} operation post-creation to verify outputs before metric computation. All other tasks use the agent's final output directly for evaluation.

\vspace{-0.2cm}
\subsection{Implementation Details}
\label{sec:implementation}
\vspace{-0.2cm}
We generate 500–1000 training tasks per benchmark using GPT-4o and train Qwen3-8B using SFT, DPO, and Agentic GRPO on A100/H200 GPUs. All models including ToolAce and xLAM are evaluated using the ReAct-based agentic pipeline with 128k context length. Full implementation details are provided in Appendix~\ref{app:implementation}.

\vspace{-0.2cm}
\section{Results and Analysis}
\vspace{-0.2cm}

This section demonstrates the effectiveness of our platform in enabling EnterpriseArena by generating high quality synthetic training data and training a specialized agent. The adaptability of the platform is further showcased by employing it across additional agentic environments.
We train a compact 8B parameter model on the generated dataset and evaluate its performance against proprietary foundation models such as GPT-4o, Claude, and Gemini, as well as leading open source tool-calling models trained on significantly larger datasets. Our evaluation spans diverse tool environments, including enterprise workflows and customer engagement scenarios, covering interaction patterns ranging from static to dynamic contexts.
We show that our platform's high-quality synthetic data enables extreme data efficiency: 8B models trained on hundreds of examples achieve competitive performance compared to models trained on thousands from existing datasets.

\vspace{-0.2cm}
\subsection{Our Platform Evaluation}
\vspace{-0.2cm}
\paragraph{Performance Comparison Against Baselines.} 
Table~\ref{tab:exec_results} reports execution accuracy of our Qwen3-8B models across four agentic environments, comparing against baselines including Qwen3-8B Base, ToolAce, xLAM-2-70B, and proprietary models: GPT-4o, Claude-3.5-Sonnet, Gemini-2.5 Pro. For benchmarks with tool-level annotations, specifically EnterpriseArena and EnterpriseBench, Table~\ref{tab:tool_results} reports tool selection accuracy. 
Qwen3-8B SFT trained on under 1K examples from our platform surpasses Qwen3-8B Base across all benchmarks and beats ToolAce and xLAM on $\tau$-Bench and CRM despite using 26-60$\times$ less data. Our agentic GRPO variant achieves substantial gains, outperforming all open-source baselines and surpassing GPT-4o on two benchmarks while approaching proprietary performance in both execution and tool selection.
These results validate our platform's ability to generate high-quality synthetic data and demonstrate its versatility across EnterpriseArena and other agentic benchmarks, enabling extreme data efficiency.

\begin{table*}[t]
\centering
\begin{minipage}[t]{0.58\textwidth}
\centering
\caption{Execution accuracy across benchmarks. EA: EnterpriseArena (Ours), EB: EnterpriseBench~\cite{vishwakarma2025can}, CRM: CRMArena~\cite{huang2025crmarena}, $\tau$-B: $\tau$-Bench~\cite{yao2025tau}. }
\label{tab:exec_results}
\small
\begin{tabular}{l|cccc}
\toprule
\textbf{Model} & \textbf{EA} & \textbf{EB} & \textbf{CRM} & \boldmath$\tau$\textbf{-B} \\
\midrule
\multicolumn{5}{c}{\textit{Closed-Source Models}} \\
\midrule
\rowcolor{GrayRow}
GPT-4o (2-shot) & 0.45 & 0.47 & 0.32 & 0.54 \\
\rowcolor{GrayRow}
Claude-3.5-Sonnet (2-shot) & 0.60 & 0.55 & 0.34 & 0.56 \\
\rowcolor{GrayRow}
Gemini-2.5 Pro (2-shot) & 0.71 & 0.55 & 0.49 & 0.59 \\
\midrule
\multicolumn{5}{c}{\textit{Open-Source Models}} \\
\midrule
\rowcolor{OpenSourceRow}
Qwen3-8B Base (2-shot) & 0.31 & 0.35 & 0.25 & 0.33 \\
\rowcolor{OpenSourceRow}
ToolAce (26K-trained) & 0.39 & 0.41 & 0.10 & 0.15 \\
\rowcolor{OpenSourceRow}
xLAM-2-70B (60K-trained) & 0.15 & 0.40 & 0.12 & 0.17 \\[2pt]
\rowcolor{OpenSourceRow}
\multicolumn{5}{c}{\underline{\textit{Our Platform-Trained Models ($<$1K)}}} \\[2pt]
\rowcolor{OpenSourceRow}
Qwen3-8B SFT & 0.35 & 0.38 & 0.30 & 0.36 \\
\rowcolor{OpenSourceRow}
\textbf{Qwen3-8B Agentic GRPO} & \textbf{0.43} & \textbf{0.51} & \textbf{0.35} & \textbf{0.42} \\
\bottomrule
\end{tabular}
\end{minipage}%
\hfill
\begin{minipage}[t]{0.38\textwidth}
\centering
\caption{Tool selection accuracy for benchmarks that provide tool-level evaluation. EA: EnterpriseArena, EB: EnterpriseBench.}
\label{tab:tool_results}
\small
\begin{tabular}{l|cc}
\toprule
\textbf{Model} & \textbf{EA} & \textbf{EB} \\
\midrule
\multicolumn{3}{c}{\textit{Closed-Source Models}} \\
\midrule
\rowcolor{GrayRow}
GPT-4o (2-shot) & 0.31 & 0.21 \\
\rowcolor{GrayRow}
Claude-3.5-Sonnet (2-shot) & 0.43 & 0.22 \\
\rowcolor{GrayRow}
Gemini-2.5 Pro (2-shot) & 0.45 & 0.24 \\
\midrule
\multicolumn{3}{c}{\textit{Open-Source Models}} \\
\midrule
\rowcolor{OpenSourceRow}
Qwen3-8B Base (2-shot) & 0.14 & 0.14 \\
\rowcolor{OpenSourceRow}
ToolAce (26K-trained) & 0.15 & 0.11 \\
\rowcolor{OpenSourceRow}
xLAM-2-70B (60K-trained) & 0.10 & 0.12 \\[2pt]
\rowcolor{OpenSourceRow}
\multicolumn{3}{c}{\underline{\textit{Our Platform-Trained Models ($<$1K)}}} \\[2pt]
\rowcolor{OpenSourceRow}
Qwen3-8B SFT & 0.20 & 0.17 \\
\rowcolor{OpenSourceRow}
\textbf{Qwen3-8B Agentic GRPO} & \textbf{0.28} & \textbf{0.21} \\
\bottomrule
\end{tabular}
\end{minipage}
\end{table*}

\vspace{-0.2cm}
\paragraph{Cost Efficiency and Practical Advantages.}
Beyond performance gains, our approach offers significant practical advantages for enterprise deployment. While achieving competitive execution accuracy, the self hosted Qwen3~8B Agentic GRPO model incurs an inference cost of approximately \$0.50--\$1.00 per million tokens, compared to proprietary foundation models such as GPT-4o, Claude-3.5-Sonnet, and Gemini-2.5-Pro accessed via AWS Bedrock, which charge approximately \$3.00 per million input tokens and up to \$15.00 per million output tokens (details in Table~\ref{tab:cost_appendix}). This corresponds to an approximately $8\times$--$10\times$ reduction in inference cost, making the approach well suited for cost sensitive, large scale deployments. Combined with the integrated capabilities of our platform for data generation, model training, trajectory collection, and evaluation, organizations can develop and deploy high quality agentic systems with greater control over data.

\vspace{-0.2cm}
\paragraph{Impact of Trajectory-Level Optimization.}
We compare Agentic GRPO against standard token-level GRPO, using a reward function adapted from Agentic GRPO and tuned for the token-level setting, and against preference-based DPO on EnterpriseBench. Agentic GRPO improves execution accuracy by approximately 10\% over token-level GRPO and 15\% over DPO, while improving tool selection accuracy by approximately 10\% over both baselines. These results indicate that trajectory-level optimization is critical for multi-turn agentic tasks, validating our platform design for collecting and training on complete agent trajectories.


\vspace{-0.2cm}
\subsection{Ablation Studies}
\vspace{-0.2cm}
\paragraph{Training Data Scale.}
We study the impact of our platform's training data quantity by training Qwen3~8B models on 500, 1000, and 1500 instances from EnterpriseBench. Increasing the training set from 500 to 1000 samples results in an approximately 2.5\% improvement in supervised training performance, while further scaling to 1500 samples leads to a performance drop of around 2\%. These diminishing returns suggest that compact models can effectively adapt to agentic tool environments using relatively small amounts of high quality data, with performance saturating beyond 1000 samples. This plateau highlights the high information density of our platform, where task diversity maintained during tool-based graph construction saturates beyond 1000 samples, as additional instances introduce redundancy within the fixed tool interaction patterns.

\vspace{-0.2cm}
\paragraph{Adaptation to Environment Changes.}
To evaluate robustness to environment changes, we introduce environment modifications to EnterpriseBench including API schema changes, new tool additions, and data modifications.These changes affect 30\% of the tool set (measured by the fraction of tools with modified schemas, parameters, or data). Table~\ref{tab:env_adaptation} shows that the original environment trained model experiences a performance drop of 15\% when evaluated on the modified environment. However, generating 200 additional trajectories under the modified environment and performing incremental training recovers substantial performance, achieving 95\% of the original accuracy with minimal additional data. This demonstrates our platform's ability to support rapid model adaptation to evolving enterprise environments without requiring full retraining.

\begin{table}[t]
\centering
\caption{Adaptation to environment changes on EnterpriseBench. Model performance before and after API schema modifications, with and without incremental training.}
\label{tab:env_adaptation}
\small
\begin{tabular}{l|cc}
\toprule
\textbf{Model} & \textbf{LLM} & \textbf{Tool} \\
& \textbf{Eval} & \textbf{Eval} \\
\midrule
Qwen3-8B AGRPO (original env) & 0.50 & 0.20 \\
Qwen3-8B AGRPO (modified env) & 0.43 & 0.15 \\
\quad + 200 samples incremental training & 0.48 & 0.18 \\
\bottomrule
\end{tabular}
\end{table}

\vspace{-0.2cm}
\subsection{Synthesized Tasks Analysis}
\vspace{-0.2cm}
We evaluate the quality of data generated by our platform along three key dimensions commonly used in synthetic data assessment: diversity, complexity, and correctness, following established evaluation works~\cite{liutoolace, havrilla2024surveying}. Our analysis is conducted on 1,500 synthetic trajectories generated for the EnterpriseBench environment.

\textbf{Diversity.} 
We measure diversity using Self-BLEU scores~\cite{zhu2018texygen}, achieving 0.4, indicating moderate diversity with acceptable variety in generated conversations. Our synthetic data pool contains 70 unique APIs spanning 5 enterprise domains with balanced distribution across categories: Software Engineering (34.4\%), Customer Relationship Management (25.3\%), Human Resources (20.8\%), General Operations (16.0\%), and IT Operations (3.5\%). This broad coverage ensures model training across diverse enterprise functions while preventing domain overfitting.
\\
\noindent \textbf{Complexity.}
Our generated tasks average 3.2 turns per dialog with standard deviation of 1.29, spanning from single-turn queries to complex multi-step workflows. Notably, 68.1\% of tasks require multi-turn reasoning and 54.7\% involve multi-tool composition with dependency chains between API calls. This complexity distribution reflects real-world enterprise scenarios, enabling effective decision-making and tool orchestration training.
\\
\noindent \textbf{Correctness.}
Following ToolACE~\cite{liutoolace}, we implement dual-layer verification. Rule-based validation checks API schema compliance, parameter type matching, and required field presence across all generated samples, achieving 100\% pass rate through schema-guided generation. Model-based semantic evaluation using GPT-4 as judge on 200 stratified samples yields 81.9\% pass rate. These results validate that our platform generates high-quality training data with both syntactic and semantic correctness.

\vspace{-0.2cm}
\subsection{Error Analysis}
\vspace{-0.2cm}
To identify systematic failure modes and guide future improvements, we analyze 50 failure cases from the Qwen3-8B model trained with Agentic GRPO on EnterpriseArena. The analysis reveals limitations related to data coverage, training signals, and model capabilities in complex multi-turn agentic reasoning.
\noindent
\begin{itemize}[noitemsep, wide=0pt, topsep=0pt]
    \item \textbf{Domain Misselection and Recursion Loops (28\%):} 
    In tasks with underspecified or ambiguous domain cues, the model frequently invokes tools from incorrect applications (e.g., selecting GitHub tools for HR-related tasks), leading to invalid tool calls and recursion-limit failures. This indicates limited robustness to domain ambiguity, as the training data predominantly contains well-specified tasks.

    \item \textbf{Tool Parameter Errors (42\%):} 
    The most common failure mode arises from incorrect tool arguments, resulting in API execution errors. Unlike larger proprietary models like Claude, Gemini that often recover through self-correction after tool failures, the local model struggles to revise invalid parameters, highlighting limitations in error recovery and tool grounding under constrained model capacity.

    \item \textbf{Task Decomposition Failures (18\%):} 
    For multi-step tasks, the model sometimes completes only the initial subtask and fails to plan or execute subsequent steps, suggesting insufficient long-horizon planning and trajectory-level credit assignment.

    \item \textbf{Context Loss (12\%):} 
    In longer interactions, the model loses coherence with earlier context, leading to incorrect assumptions or premature termination. This reflects challenges in maintaining state and intent over extended tool-based dialogues.
\end{itemize}

\vspace{-0.2cm}
\section{Related Work}
\vspace{-0.2cm}

\textbf{Tool Learning and Agent Benchmarks.}
LLMs have rapidly improved at tool use and decision-making via API calling and interactive execution \cite{qintoolllm, qian2023toolalpaca, liutoolace, schick2023toolformer, qu2024exploration}. Evaluation has progressed from early single-domain tool suites (e.g., ToolBench, API-Bank) \cite{qintoolllm, li2023api} to richer agent environments spanning web navigation, software engineering, and workplace workflows (e.g., WebArena, SWE-bench, AgentCompany, TravelPlanner) \cite{zhouwebarena, yangswe, xu2024theagentcompany, xie2024travelplanner}. As environments become multi-domain and operationally constrained, smaller enterprise-oriented models often struggle to generalize across heterogeneous workflows \cite{shen2024small, manduzio2024improving}. \name{} targets this gap by enabling scalable training and evaluation of small agentic models in multi-application environments via dynamically simulated tasks and interaction trajectories.
\\
\noindent \textbf{Environment Exploration and Task Synthesis.}
Recent work scales synthetic task generation through knowledge-based generation, interaction logging, and exploration-driven synthesis, particularly for web agents \cite{ou2024synatra, lai2024autowebglm, murty2024bagel, murty2024nnetnav, gandhi2025go, sun2025genesis, ramrakhya2025scaling}. For tool-calling, related efforts generate tasks from structured assets such as knowledge graphs or API documentation \cite{chen2025graph2eval, liutoolace}. In contrast, \name{} generates tasks directly from schemas exposed by the environment itself, reducing reliance on manually curated graphs or hand-designed, task-level API abstractions.
\\
\noindent \textbf{Agent Adaptation and Enterprise Orchestration.}
Tool-use learning commonly relies on supervised fine-tuning over expert trajectories \cite{weifinetuned, qintoolllm}, while preference optimization (e.g., DPO) \cite{rafailov2023direct} and reinforcement learning variants increasingly support adaptation under distribution shift and environment feedback. Our training setup supports SFT, DPO, and Agentic GRPO \cite{singh2025agentic, shao2024deepseekmath} within the same interactive loop to study adaptation in both static and dynamic settings. Complementary to enterprise-focused evaluations such as CRMArena and $\tau$-Bench \cite{huang2025crmarena, huang2025crmarenapro, yao2025tau} and broader suites like AgentBench/WebArena/WorkArena \cite{liu2024agentbench, zhouwebarena, drouin2024workarena}, EnterpriseArena emphasizes cross-application orchestration with evolving schemas and inter-application dependencies; a tabular comparison appear in Table~\ref{tab:enterprise_comparison}.
\\
\noindent \textbf{We provide an extended related work in Appendix~\ref{app:extended_related_work}.}

\vspace{-0.2cm}
\section{Conclusion}
\vspace{-0.2cm}
In this work, we presented EnterpriseLab, a platform designed to consolidate fragmented corporate data and tool ecosystems into a cohesive environment for agent development. By integrating a modular Model Context Protocol structure with automated trajectory synthesis, the platform provides a systematic approach to bridging the gap between disjointed enterprise tools and agentic training cycles. Our empirical evaluation on EnterpriseArena, a benchmark environment instantiated via the platform, indicates that this architectural unification facilitates the generation of high-quality training data, enabling 8B-parameter models to reach performance levels comparable to larger proprietary systems. The architectural adaptability of the platform is further evidenced by its performance across external benchmarks, demonstrating that a unified, modular infrastructure supports the development of logic that remains robust across diverse corporate environments. Ultimately, this research indicates that the architectural unification of fragmented enterprise ecosystems plays a central role in enabling agentic capability, providing a scalable pathway for developing specialized enterprise models through an automated, environment-aligned pipeline.

\vspace{-0.2cm}
\section*{Impact Statement}
\label{sec:impact}
\vspace{-0.2cm}
This work aims to democratize access to enterprise AI agents by enabling organizations to train high-performance models that are cost-effective and privacy-preserving. The primary benefits include improved accessibility for smaller organizations without dependence on expensive proprietary APIs, enhanced data privacy through on-premise deployment that addresses compliance requirements, an 8-10× reduction in inference costs making AI agents economically viable, and reduced computational requirements that lower energy consumption compared to frontier models. There are few societal consequences of our work, none of which we feel must be specifically highlighted here.

\bibliography{example_paper}
\bibliographystyle{icml2026}

\newpage
\appendix
\onecolumn

\section{Appendix}

\noindent This appendix provides supplementary material that supports the main paper. Due to space constraints, we defer implementation details, extended results, algorithms, and design choices to this section. Specifically, we include additional experimental results, notation reference, detailed reward formulations, extended related work, expert study protocols, MCP server tool specifications, and all the prompts. The appendix is organized as follows:

\begin{enumerate}[nosep]
    \item \hyperref[app:add_res]{Additional Results, EnterpriseArena Details, and Supplementary Details}
    \item \hyperref[app:algo]{Algorithm of Agentic GRPO with ReAct Sampling}
    \item \hyperref[app:reward]{Trajectory Reward Design}
    \item \hyperref[app:extended_related_work]{Extended Related Work}
    \item \hyperref[app:implementation]{Implementation Details}
    \item \hyperref[app:exp_study]{Expert Study Details}
    \item \hyperref[app:mcp_desc]{MCP Servers Information}
    \item \hyperref[app:lim]{Limitations}
    \item \hyperref[app:prompt]{Prompts}
\end{enumerate}

\subsection{Additional Results, EnterpriseArena Details, and Supplementary Details}
\label{app:add_res}

Additional Results. We conduct additional experiments using Qwen2.5-4B trained with our platform on EnterpriseBench to demonstrate the platform's generalizability across different model architectures. Table~\ref{tab:enterprisebench_gpt4o} reports results when GPT-4o is used as the evaluator. The trained model achieves approximately 20\% improvement over the base model, with Agentic GRPO enabling the 4B model to reach GPT-4o-level performance.

Table~\ref{tab:enterprisebench_claude} presents EnterpriseBench results evaluated using Claude-4.5-Sonnet.
Across both evaluation settings, GPT-4o and Claude-4.5-Sonnet, we observe only minor differences in performance. This consistency indicates that the reported results are stable with respect to the choice of evaluator model, largely because the evaluation rubrics are concrete and execution focused rather than subjective.

Table~\ref{tab:cost_appendix} provides inference costs for proprietary models accessed via AWS Bedrock and our self hosted Qwen3-8B Agentic GRPO model, based on publicly available pricing as of January 2026.\footnote{\url{https://aws.amazon.com/bedrock/pricing/}}

\begin{table}[H]
\centering
\caption{EnterpriseBench results evaluated by GPT-4o. Scores represent execution scores on enterprise task completion.}
\label{tab:enterprisebench_gpt4o}
\begin{tabular}{lc}
\toprule
\textbf{Model} & \textbf{Score} \\
\midrule
GPT-4o & 0.47 \\
ToolAce & 0.41 \\
XLAM-2-70B & 0.40 \\
\midrule
Qwen3-4B (Base) & 0.27 \\
Qwen3-4B (SFT) & 0.32 \\
Qwen3-4B (Agentic GRPO) & 0.38 \\
\bottomrule
\end{tabular}
\end{table}


\begin{table}[H]
\centering
\caption{EnterpriseBench results evaluated by Claude-4.5-Sonnet. Scores represent execution scores on enterprise task completion.}
\label{tab:enterprisebench_claude}
\begin{tabular}{lc}
\toprule
\textbf{Model} & \textbf{Score} \\
\midrule
GPT-4o & 0.44 \\
ToolAce & 0.39 \\
XLAM-2-70B & 0.39 \\
\midrule
Qwen3-4B (Base) & 0.25 \\
Qwen3-4B (SFT) & 0.31 \\
Qwen3-4B (Agentic GRPO) & 0.36 \\
\bottomrule
\end{tabular}
\end{table}


\begin{table}[H]
\centering
\small
\begin{tabular}{l|cc}
\toprule
\textbf{Model} & \textbf{Input} & \textbf{Output} \\
& \textbf{(\$ per 1M tokens)} & \textbf{(\$ per 1M tokens)} \\
\midrule
GPT-4o & 5.00 & 15.00 \\
Claude-3.5-Sonnet & 3.00 & 15.00 \\
Gemini-2.5 Pro & 1.25 & 10.00 \\
\midrule
Qwen3-8B Agentic GRPO & \multicolumn{2}{c}{0.50--1.00} \\
\bottomrule
\end{tabular}
\caption{Inference costs via AWS Bedrock for proprietary models and self-hosted deployment for our model on AWS g5.2xlarge instances.}
\label{tab:cost_appendix}
\end{table}


\textbf{EnterpriseArena Details.} 
In Table~\ref{tab:enterprise_comparison} compares EnterpriseArena with existing agent benchmarks, highlighting differences in domain coverage, multi-application orchestration, and data dynamics. Table~\ref{tab:domain_dist} shows the task distribution across enterprise domains and workflow categories. Table~\ref{tab:complex_task_examples} provides representative examples of complex multi-step tasks, showcasing cross-functional orchestration patterns and specific tool sequences across software engineering, HR, CRM, and project management domains.

\begin{table}[H]
    \centering
    \caption{Comparison of EnterpriseArena with existing agent benchmarks. EnterpriseArena uniquely targets multi-application enterprise orchestration with dynamic data, distinguishing it from single-domain (CRM, Code) or static benchmarks.}
    \label{tab:enterprise_comparison}
    \resizebox{\textwidth}{!}{
    \begin{tabular}{lcccccc}
        \toprule
        \textbf{Benchmark} & \textbf{Domain Focus} & \textbf{Multi-App Flow} & \textbf{Dynamic Data} & \textbf{Enterprise Constraints} & \textbf{Key Limitation} \\
        \midrule
        AgentBench \cite{liu2024agentbench} & General Reasoning & \ding{55} & \ding{55} & \ding{55} & No Enterprise Context \\
        WebArena \cite{zhouwebarena} & General Web UI & \ding{55} & \ding{51} (Live Sites) & \ding{55} & Public Web, No Authorization \\
        SWE-bench \cite{yangswe} & Software Eng. & \ding{55} (GitHub only) & \ding{55} (Repo Snapshot) & \ding{55} & Coding Domain Only \\
        WorkArena \cite{drouin2024workarena} & IT/ServiceNow & \ding{55} (Single Platform) & \ding{51} (Live UI) & \ding{51} & Single SaaS Ecosystem \\
        CRMArena \cite{huang2025crmarena} & CRM & \ding{55} (Salesforce) & \ding{55} (Static) & \ding{51} & Single Domain (CRM) \\
        Tau-bench \cite{yao2025tau} & Customer Service & \ding{55} & \ding{51} (Stateful) & \ding{55} & Limited Tool Diversity \\
        EnterpriseBench \cite{vishwakarma2025can} & General Enterprise & \ding{51} & \ding{55} (Static DB) & \ding{51} & Static Data \& Functions \\
        AgentCompany \cite{xu2024theagentcompany} & Corporate Admin & \ding{51} (Simulated OS) & \ding{55} & \ding{51} & Requires Interface Interaction \\
        \midrule
        \textbf{\textsc{EnterpriseArena} (Ours)} & \textbf{Cross-Functional} & \textbf{\ding{51} (CRM, HR, Finance)} & \textbf{\ding{51}} & \textbf{\ding{51} } & \textbf{-} \\
        \bottomrule
    \end{tabular}
    }
\end{table}

\begin{table}[H]
\centering
\small
\caption{Examples of complex multi-step tasks in EnterpriseArena demonstrating cross-functional orchestration with specific tool sequences.}
\label{tab:complex_task_examples}
\begin{tabular}{p{2.5cm} p{5.5cm} p{2.5cm} p{3.0cm}}
\toprule
\textbf{Domain} & \textbf{Task Description} & \textbf{Task Category} & \textbf{Tools Used} \\
\midrule

Software Engineering & Create a project named 'Nexus', add frontend, backend, database modules, and create issues in each module. & Multi-step CRUD & \texttt{create\_project}, \texttt{create\_module} (×3), \texttt{create\_issue} (×3), \texttt{send\_channel\_message} \\

\midrule

HR \& Recruitment & Read job description from Documents, fetch shortlisted resumes, identify top candidates, and coordinate interview scheduling. & Search \& Orchestration & \texttt{list\_files}, \texttt{read\_file\_content}, \texttt{search\_files}, \texttt{get\_employees}, \texttt{send\_email} (×2) \\

\midrule

IT \& Project Mgmt & Announce MCP agent online in general channel, create MCP-adapters issue in Sprint 2, verify it appears in module. & Communication \& CRUD & \texttt{send\_channel\_message}, \texttt{create\_issue}, \texttt{list\_issues}, \texttt{create\_cycle} \\

\midrule

CRM \& Business & Create customer TechCorp, add three products to catalog, generate sales order, create invoice, email confirmation. & Multi-entity Workflow & \texttt{create\_customer}, \texttt{create\_product} (×3), \texttt{create\_order}, \texttt{create\_invoice}, \texttt{send\_email} \\

\midrule

DevOps & Search for repository, fork to personal namespace, create feature branch, push config files, open merge request. & Version Control & \texttt{search\_repositories}, \texttt{fork\_repository}, \texttt{create\_branch}, \texttt{push\_files}, \texttt{create\_merge\_request} \\

\midrule

File \& Storage & List project documents, create Finance folder in OwnCloud, upload monthly reports, share folder info via channel. & Data Organization & \texttt{list\_files}, \texttt{create\_folder}, \texttt{upload\_file} (×2), \texttt{get\_storage\_info}, \texttt{send\_channel\_message} \\

\midrule

Multi-Department & Create holiday request for employee, retrieve timesheets, upload to Documents folder, notify manager. & Cross-functional & \texttt{create\_holiday\_request}, \texttt{get\_timesheets}, \texttt{upload\_file}, \texttt{get\_employee}, \texttt{send\_user\_message} \\

\midrule

IT Ticketing & List urgent tickets assigned to me, search for payment gateway issues, update priority to critical, assign to agent. & Search \& Update & \texttt{search\_tickets}, \texttt{get\_ticket} (×2), \texttt{update\_ticket} (×2), \texttt{assign\_agent} \\

\bottomrule
\end{tabular}
\end{table}

\begin{table}[h]
\centering
\small
\caption{Task distribution across enterprise domains and workflow categories.}
\label{tab:domain_dist}
\begin{tabular}{lc|lc}
\toprule
\textbf{Domain} & \textbf{\%} & \textbf{Category} & \textbf{\%} \\
\midrule
IT \& Project Mgmt & 22 & CRUD Operations & 35 \\
Software Engineering & 18 & Search \& Orchestration & 28 \\
Human Resources & 15 & Multi-entity Workflow & 18 \\
CRM \& Business & 14 & Version Control & 12 \\
DevOps & 12 & Cross-functional & 7 \\
Communication & 10 & & \\
Finance \& Payroll & 6 & & \\
Multi-Department & 3 & & \\
\bottomrule
\end{tabular}
\end{table}


\noindent \textbf{Additional Details.}
Table~\ref{tab:experiment_data} summarizes the train and test split sizes used for all evaluated benchmarks.

\begin{table}[H]
\centering
\caption{Train/Test size for Experimental setting}
\label{tab:experiment_data}
\begin{tabular}{l c c}
\toprule
\textbf{Benchmark} & \textbf{Train (Size)} & \textbf{Test (Size)} \\
\midrule
EnterpriseArena    & 500                & 500 \\
EnterpriseBench    & 700                 & 500 \\
CRMArena           & 600                   & 500 (Function Calling) \\
Tau-Bench          & 300 (Retail), 300 (Airline)                   & 115 (Retail), 50 (Airline) \\
\bottomrule
\end{tabular}
\end{table}



\subsection{Algorithm of Agentic GRPO with ReAct Sampling}
\label{app:algo}
This algorithm trains an agent using Group Relative Policy Optimization with interleaved ReAct trajectories. For each query, it samples G trajectories by alternating between reasoning and tool-based actions, computes group-normalized advantages from multi-component rewards, and updates the policy using a KL-regularized objective.
Table~\ref{tab:notation} defines the notation used throughout our Agentic GRPO formulation and algorithm descriptions.

\begin{algorithm}[t]
\small
\caption{Agentic GRPO with ReAct Sampling}
\label{alg:agentic_grpo}
\begin{algorithmic}[1]
\REQUIRE Policy $\pi_\theta$, Ref $\pi_{\mathrm{ref}}$, Dataset $\mathcal{D}$, Group $G$
\STATE \textbf{Notation:} All symbols and variables are defined in Table~\ref{tab:notation}.
\FOR{each training iteration}
    \STATE Sample batch $Q \sim \mathcal{D}$
    \FOR{each $q \in Q$}
        \STATE $\mathcal{T}_q \leftarrow \emptyset$
        \FOR{$g = 1$ to $G$}
            \STATE $h \leftarrow q$, $steps \leftarrow 0$
            \WHILE{$steps < T_{\max}$}
                \STATE $z \sim \pi_\theta(\cdot \mid h)$
                \STATE $a \sim \pi_\theta(\cdot \mid h, z)$
                \IF{$a$ is tool call}
                    \STATE $o \leftarrow \mathrm{Env}(a)$
                    \STATE $h \leftarrow h \oplus z \oplus a \oplus o$
                \ELSE
                    \STATE $h \leftarrow h \oplus z \oplus a$
                    \STATE \textbf{break}
                \ENDIF
                \STATE $steps \leftarrow steps + 1$
            \ENDWHILE
            \STATE Add trajectory $\tau_g \leftarrow h$ to $\mathcal{T}_q$
        \ENDFOR
        \FOR{each $\tau_g \in \mathcal{T}_q$}
            \STATE $r_g = \sum_{k=1}^{4} w_k r_k(\tau_g)$
        \ENDFOR
        \STATE $\hat{A}_g = \frac{r_g - \mathrm{mean}(r_{1:G})}{\mathrm{std}(r_{1:G}) + \epsilon}$ for all $g$
    \ENDFOR
    \STATE $\mathcal{L} = -\frac{1}{|Q|} \sum_{q \in Q} \frac{1}{G} \sum_{g=1}^{G} \hat{A}_g \sum_{t=1}^{T_g} \log \pi_\theta(a_{g,t}\mid h_{g,<t}) + \beta D_{KL}(\pi_\theta \Vert \pi_{\mathrm{ref}})$
    \STATE Update $\theta$ using $\nabla_\theta \mathcal{L}$
\ENDFOR
\end{algorithmic}
\end{algorithm}

\begin{table}[H]
\centering
\small
\setlength{\tabcolsep}{2pt}
\begin{tabularx}{\linewidth}{@{}l X @{\hspace{1em}} l X@{}}
\toprule
\textbf{Sym.} & \textbf{Meaning} & \textbf{Sym.} & \textbf{Meaning} \\
\midrule
$\pi_\theta$ & Policy model (LLM) & $G$ & Group size (rollouts/query) \\
$\theta$ & Trainable params (LoRA) & $g$ & Trajectory index $1 \dots G$ \\
$\pi_{\text{ref}}$ & Frozen reference policy & $\mathcal{T}_q$ & Trajectory set for query $q$ \\
$\mathcal{D}$ & Training dataset & $h$ & History (msgs buffer) \\
$Q$ & Batch sampled from $\mathcal{D}$ & $T_{\max}$ & Max tool steps limit \\
$q$ & Single query/task & $z$ & Thought tokens (\texttt{<think>}) \\
$a$ & Action/Tool call & $o$ & Observation (tool output) \\
$\tau_g$ & Full trajectory transcript & $\beta$ & KL penalty coefficient\\
$r_g$ & Total trajectory reward & $\hat{A}_g$ & Group-relative advantage \\
$D_{KL}$ & KL divergence term & $\epsilon$ & Stability constant \\
\bottomrule
\end{tabularx}
\caption{Notation for Agentic GRPO}
\label{tab:notation}
\end{table}

\subsection{Trajectory Reward Design}
\label{app:reward}

We design a trajectory-level reward function $r(\tau)$ to evaluate both task completion and correct interaction with the environment. The reward is computed by executing the reference trajectory $\tau$ in the environment and aggregating multiple execution-grounded signals.

Specifically, the reward comprises the following components:
\begin{itemize}[leftmargin=*, noitemsep]
    \item \textbf{Tool selection accuracy} $r_1(\tau)$: measures whether the agent selects appropriate tools at each step of the trajectory.
    \item \textbf{Execution success} $r_2(\tau)$: verifies that all tool calls execute without runtime errors.
    \item \textbf{Final answer correctness} $r_3(\tau)$: evaluates whether the final output satisfies the task objective according to environment-defined success criteria.
    \item \textbf{Format compliance} $r_4(\tau)$: checks adherence to the required ReAct-style response format.
\end{itemize}

The overall trajectory reward is computed as a weighted sum:
\[
r(\tau) = \sum_{k=1}^{4} w_k \, r_k(\tau), \quad \text{where } \sum_k w_k = 1,
\]
and is normalized to lie in $[0,1]$. Trajectories whose reference executions fail automatically receive zero reward. This execution-grounded reward formulation avoids reliance on learned reward models and enables stable optimization of multi-step, tool-augmented behaviors.

\subsection{Extended Related Work}
\label{app:extended_related_work}
\vspace{-0.2cm}
\textbf{Tool Learning.} Recent work demonstrates LLMs' growing mastery of tools and decision-making in complex environments \cite{qintoolllm, qian2023toolalpaca, liutoolace, schick2023toolformer, qu2024exploration}. Tool-based agent environments have evolved from simple, single-domain benchmarks like ToolBench \cite{qintoolllm} and API-Bank \cite{li2023api} to complex, multi-domain ones such as WebArena \cite{zhouwebarena} (web navigation), SWE-bench \cite{yangswe} (software engineering), AgentCompany \cite{xu2024theagentcompany} (workplace workflows), and TravelPlanner \cite{xie2024travelplanner} (planning tasks). As environments grow increasingly complex, single-domain agents struggle to generalize to multi-domain workflows, particularly enterprise-constrained small models lacking proprietary LLM capacity \cite{shen2024small, manduzio2024improving}. This challenge motivates the development of \name{}, a platform that enables scalable training of small agentic models across multi-application environments through dynamic simulation of tasks and agent interaction trajectories.

\textbf{Environment Exploration.}
There have been several works on scaling synthetic task generation for different domains. For web agents, approaches include static knowledge-based generation \cite{ou2024synatra}, interaction logging \cite{lai2024autowebglm, murty2024bagel}, and exploration-based synthesis \cite{murty2024nnetnav, gandhi2025go, sun2025genesis, ramrakhya2025scaling}. For tool-calling agents, Graph2Eval \cite{chen2025graph2eval} generates tasks by sampling subgraphs from knowledge graphs, while ToolACE \cite{liutoolace} synthesizes tool-calling conversations from API documentation. However, these approaches require predefined knowledge graphs or API schemas, limiting their applicability to new tool environments. \name{} addresses this by generating tasks directly from environment-exposed schemas, avoiding reliance on manually curated knowledge graphs or manually designed, task-level API abstractions.

\textbf{Agent Adaptation.} 
As deployment costs and environmental concerns grow, the pursuit of efficient language models has intensified, with recent work showing that smaller models trained on high-quality data can match or even surpass much larger counterparts \cite{touvron2023llama}. Supervised fine-tuning (SFT) is the standard approach for learning tool-use patterns from expert trajectories \cite{weifinetuned, qintoolllm}, but it remains vulnerable to distribution shift as workflows evolve over time. To operate effectively in specialized and dynamic environments, agent training has therefore shifted toward more advanced alignment and reinforcement-learning-based strategies. For preference-based alignment in constrained decision-making settings, Direct Preference Optimization (DPO) \cite{rafailov2023direct} enables agents to learn from paired preferences. However, preference-based methods remain largely offline and do not directly incorporate environment feedback during execution. To address this, we incorporate Agentic Group Relative Policy Optimization (Agentic GRPO), a reinforcement-learning setting adapted from the ARTIST framework \cite{singh2025agentic}, in which the agent alternates between reasoning and tool execution, integrating environment interaction directly into the optimization loop, unlike standard GRPO \cite{shao2024deepseekmath}. \name{} supports each of these training paradigms for both static and dynamic benchmarks, providing a versatile testbed for studying agent adaptation across SFT, DPO, and Agentic GRPO.

\textbf{Environments for Agentic Orchestration.} 
Evaluating agents in realistic business workflows has gained traction with specialized benchmarks such as CRMArena \cite{huang2025crmarena, huang2025crmarenapro} and $\tau$-Bench \cite{yao2025tau}, which assess agents in CRM and customer service domains. Broader evaluation suites including AgentBench \cite{liu2024agentbench}, WebArena \cite{zhouwebarena}, and WorkArena \cite{drouin2024workarena} evaluate general reasoning and web UI navigation capabilities, but often lack interconnected operational data and cross-application dependencies. Similarly, AgentCompany \cite{xu2024theagentcompany} and EnterpriseBench \cite{vishwakarma2025can} simulate corporate environments with long-horizon tasks, yet typically rely on static task definitions or simulated OS interactions, rather than modular SaaS-style applications with evolving schemas and inter-application dependencies. Specialized benchmarks such as SWE-bench \cite{yangswe} focus deeply on software engineering workflows, but do not generalize to cross-functional enterprise operations.
Addressing these limitations, we introduce EnterpriseArena, a platform designed for agentic orchestration in dynamic, multi-tool enterprise environments. EnterpriseArena leverages \name{}’s scalable environment generation to enable workflow execution across heterogeneous applications (e.g., CRM, HR, Finance), incorporating enterprise-style access controls and cross-application data dependencies.
 Table~\ref{tab:enterprise_comparison} compares EnterpriseArena with existing benchmarks, highlighting its unique support for multi-application orchestration and dynamic tool interaction.

\subsection{Implementation Details}
\label{app:implementation}
\vspace{-0.2cm}
\textbf{Task Generation.} We use GPT-4o\footref{gpt4o} with the tasks synthesis pipeline 
to generate 500 to 1000 training tasks per benchmark at temperature 0.7.

\noindent\textbf{Training Configuration.} 
We train Qwen3-8B on $2\times$A100 GPUs (80GB each) for SFT and DPO, and $4\times$H200 GPUs (140GB each) for Agentic GRPO. 
SFT uses LoRA targeting \texttt{q\_proj}, \texttt{k\_proj}, \texttt{v\_proj}, \texttt{o\_proj} (rank 128, $\alpha=256$, lr $5\times10^{-5}$, batch size 4, weight decay 0.005) for 2 epochs. 
DPO applies the same LoRA configuration on preference pairs from base and SFT model trajectories.
Agentic GRPO uses group size $G=4$ (4 trajectories per task) with learning rate $1\times10^{-5}$ during online training.
Training time ranges from 30 minutes to 2 hours for SFT and DPO, and 24-30 hours for 
Agentic GRPO.

\noindent\textbf{Inference Configuration.}
For proprietary models including GPT-4o, Claude-3.5-Sonnet, and Gemini-2.5-Pro, we use a temperature of 0.7, top-$p$ of 0.95, and a maximum output length of 16k tokens, with 2-shot prompting using manually created and verified task examples.
All models, including proprietary and open source models such as xLAM-2-70B, ToolACE, and our platform trained models are evaluated using the ReAct-based agentic pipeline. Models alternate between generating reasoning, issuing tool calls to the environment backend, receiving observations, and producing subsequent actions. The maximum input context length is fixed to 128k tokens for all models.

\subsection{Expert Study Details}
\label{app:exp_study}

\begin{figure*}[htbp]
    \centering
    \begin{subfigure}[b]{0.45\textwidth}
        \centering
        \includegraphics[height=6.0cm]{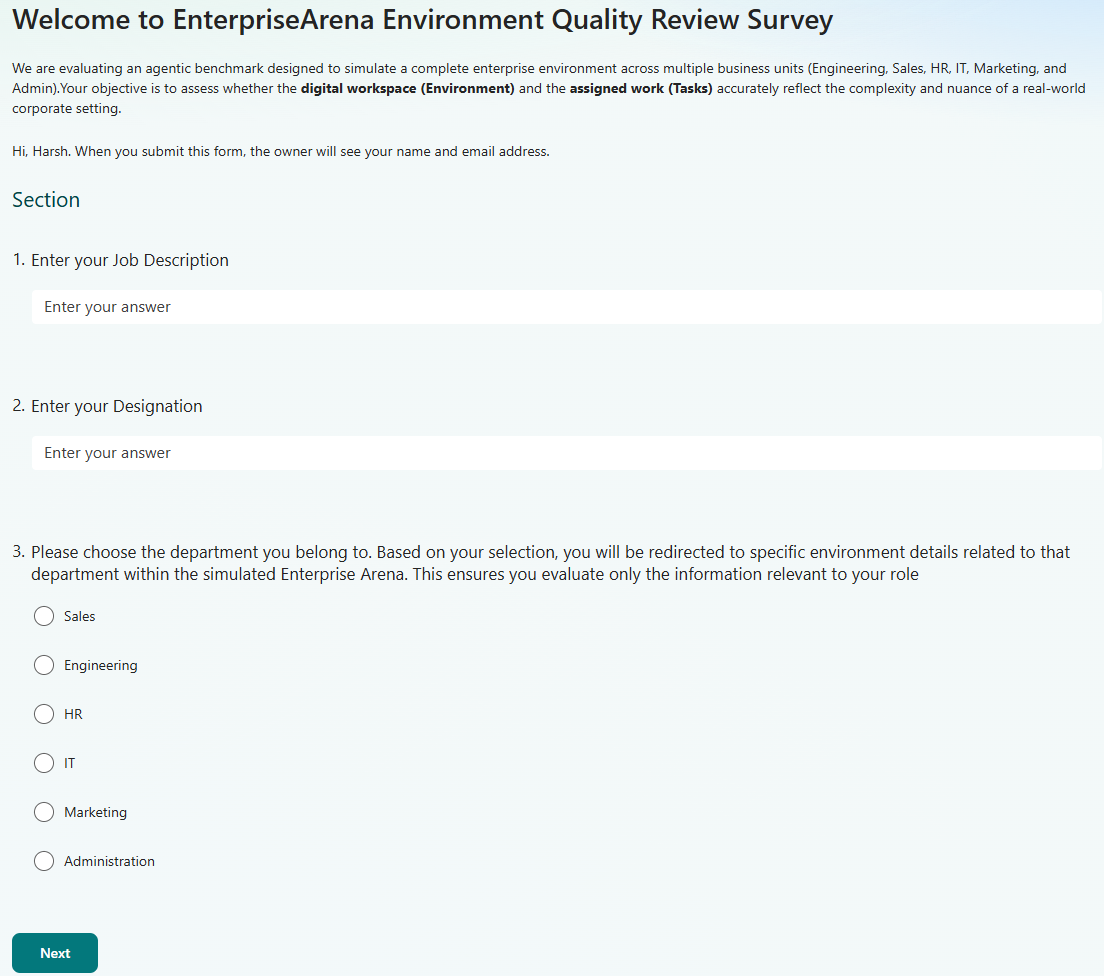}
        \caption{First page of the Microsoft Form used to collect information about domain experts, including their department and position.}
        \label{fig:img1}
    \end{subfigure}
    \hfill
    \begin{subfigure}[b]{0.60\textwidth}
        \centering
        \includegraphics[height=4.4cm]{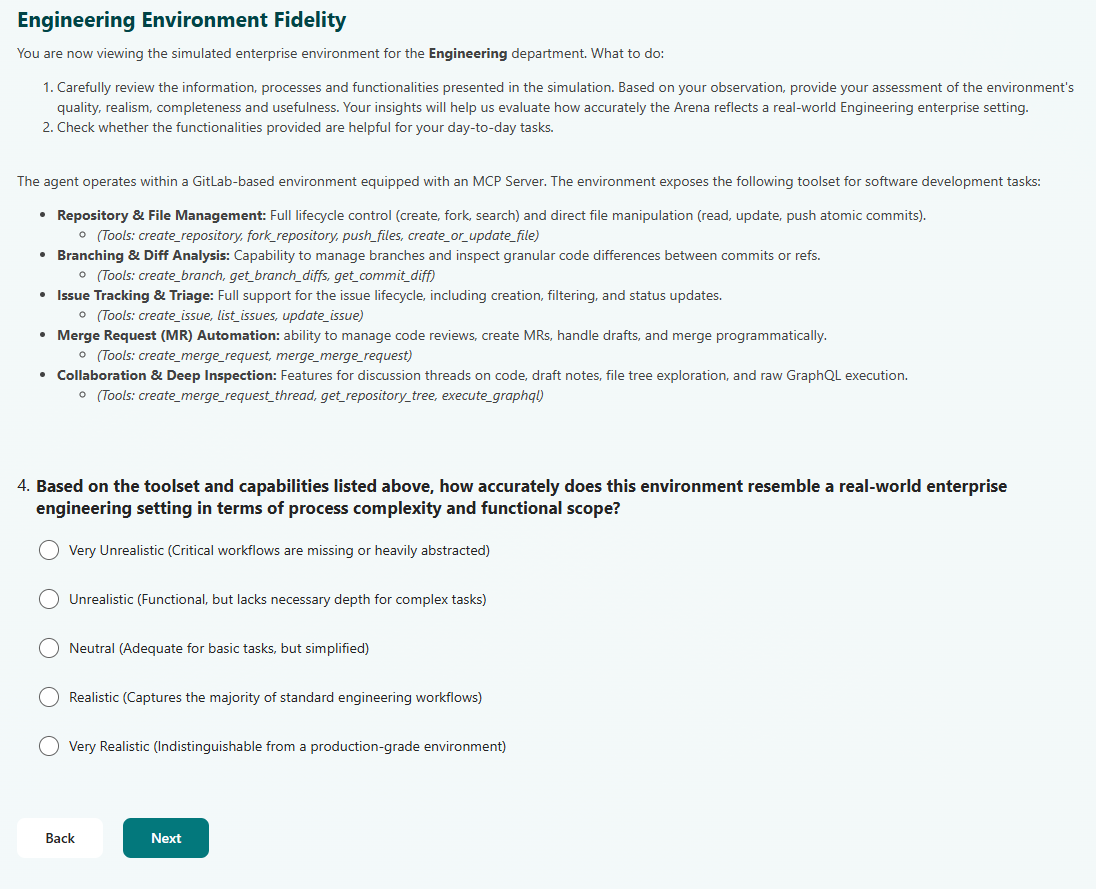}
        \caption{Next page of the form displaying simulated environment details for the selected department. Users are asked to rate the realism of the functionalities provided for the selected department, choosing from options ranging from `Very Unrealistic' to `Very Realistic.' They also have to provide reasons when selecting `Unrealistic'.}
        \label{fig:img2}
    \end{subfigure}

    \vspace{0.5cm}

    \begin{subfigure}[b]{0.45\textwidth}
        \centering
        \includegraphics[height=6.0cm]{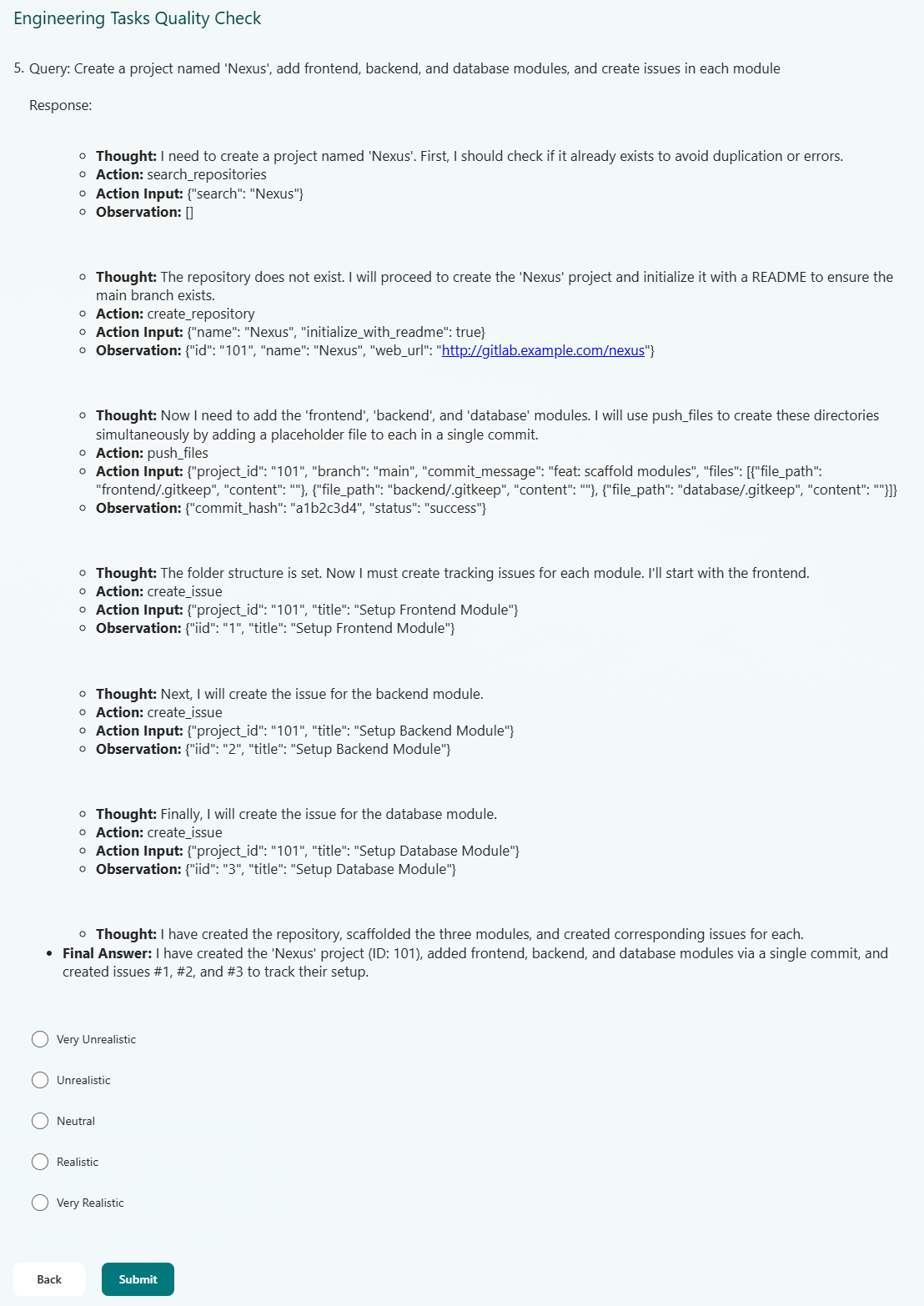}
        \caption{This page presents enterprise tasks for evaluation. Users rate each task's realism from 'Very Unrealistic' to 'Very Realistic,' and provide reasons if they select `Neutral,' `Unrealistic,' or `Very Unrealistic'.}
        \label{fig:img3}
    \end{subfigure}
    \hfill

  \caption[\textbf{Domain Expert Validation for Realism of EnterpriseArena Tasks}]{\textbf{Domain Expert Validation in EnterpriseArena.} Domain experts from all benchmark domains evaluate the fidelity of the created environment and tasks. This example shows screenshots of MS form for different steps a domain expert completes during the validation process.}

    \label{fig:cluster}
    \vspace{-0.4cm}
\end{figure*}

To validate the realism of EnterpriseArena tasks and environment specifications, we conducted an expert study with nine professionals across Software Engineering, Business Development, Sales, IT Security, Human Resources, and Finance (Table~\ref{tab:profession_data}). Using a structured Microsoft Form (Figure~\ref{fig:cluster}), experts rated the realism of MCP server functionalities and domain-specific tasks on a five-point Likert scale. Tasks rated ``Realistic'' or above were retained in the final benchmark, while lower-rated tasks were revised or discarded based on expert feedback, ensuring the 500 tasks in EnterpriseArena reflect authentic enterprise workflows.

\begin{table}[H]
\centering
\caption{\textbf{Domain Expert Demographics:} Professional roles, gender, and age distribution of the nine domain experts who validated EnterpriseArena tasks and environment specifications.}
\label{tab:profession_data}
\begin{tabular}{lcc}
\toprule
\textbf{Profession} & \textbf{Gender} & \textbf{Age} \\
\midrule
Software Engineer & M & 25 \\
Senior Engineer & F & 29 \\
\midrule
Business Development Representative & F & 26 \\
Sales Manager & F & 35 \\
\midrule
IT Engineer & M & 29 \\
IT Security Head & M & 32 \\
\midrule
HR Head & F & 40 \\
Lead HR & F & 30 \\
\midrule
Finance Manager & F & 36 \\
\bottomrule
\end{tabular}
\end{table}

\subsection{MCP Servers Information}
\label{app:mcp_desc}

EnterpriseArena's modular architecture is built on 15 specialized MCP servers that collectively expose over 140 tools spanning enterprise workflows. Table~\ref{tab:mcp_servers} provides an overview of the MCP server ecosystem, categorized by functional domain. Each server encapsulates a specific enterprise application (e.g., GitLab for version control, Frappe HR for human resources, Dolibarr for CRM) and exposes a standardized tool interface via the Model Context Protocol. The following subsections provide comprehensive tool catalogs for all servers, detailing each tool's functionality and required parameters to facilitate benchmark reproducibility and extension.

\begin{table*}[t]
    \centering
    \small
    \caption{MCP Server ecosystem in EnterpriseArena with representative tools per server.}
    \label{tab:mcp_servers}
    \resizebox{\textwidth}{!}{
    \begin{tabular}{llc p{7cm}}
        \toprule
        \textbf{Category} & \textbf{MCP Server} & \textbf{\# Tools} & \textbf{Representative Tools} \\
        \midrule
        \multirow{2}{*}{Communication} & RocketChat & 12 & \texttt{send\_user\_message}, \texttt{send\_channel\_message}, \texttt{get\_channels} \\
        & Mail System & 8 & \texttt{send\_email}, \texttt{read\_email}, \texttt{search\_emails}, \texttt{draft\_email} \\
        \midrule
        Development & GitLab MCP & 22 & \texttt{create\_merge\_request}, \texttt{create\_issue}, \texttt{push\_files}, \texttt{execute\_graphql} \\
        \midrule
        \multirow{2}{*}{Operations \& IT} & Zammad & 10 & \texttt{create\_ticket}, \texttt{assign\_agent}, \texttt{update\_status}, \texttt{add\_note} \\
        & Plane (Jira) & 14 & \texttt{create\_issue}, \texttt{add\_milestone}, \texttt{assign\_task}, \texttt{update\_state} \\
        \midrule
        \multirow{2}{*}{Human Resources} & Frappe HR & 14 & \texttt{get\_employee}, \texttt{create\_leave\_request}, \texttt{get\_attendance}, \texttt{get\_payroll} \\
        & Calendar & 6 & \texttt{create\_event}, \texttt{update\_event}, \texttt{list\_events}, \texttt{search\_events} \\
        \midrule
        \multirow{2}{*}{Data \& Storage} & Mongoose MCP & 6 & \texttt{create\_record}, \texttt{query\_database}, \texttt{update\_employee}, \texttt{delete\_record} \\
        & OwnCloud & 9 & \texttt{upload\_file}, \texttt{download\_file}, \texttt{search\_files}, \texttt{read\_file\_content} \\
        \midrule
        \multirow{2}{*}{Business (CRM)} & Dolibarr & 11 & \texttt{create\_customer}, \texttt{create\_product}, \texttt{log\_sale}, \texttt{create\_holiday\_request} \\
        & Salesforce CRM & 8 & \texttt{create\_account}, \texttt{create\_contact}, \texttt{create\_lead}, \texttt{create\_opportunity} \\
        \midrule
        Finance & Invoice System & 7 & \texttt{create\_invoice}, \texttt{send\_invoice}, \texttt{record\_payment}, \texttt{list\_invoices} \\
        \midrule
        Collaboration & Slack & 9 & \texttt{send\_message}, \texttt{create\_channel}, \texttt{upload\_file}, \texttt{get\_history} \\
        \midrule
        \multirow{2}{*}{Utilities} & File System \& Bash & 10 & \texttt{read\_file}, \texttt{write\_file}, \texttt{execute\_command} \\
        & Browser (Playwright) & 8 & \texttt{navigate}, \texttt{click}, \texttt{type}, \texttt{take\_screenshot} \\
        \bottomrule
        \multicolumn{3}{r}{\textbf{Total (15 servers):}} & \textbf{140+ Tools} \\
    \end{tabular}
    }
\end{table*}

\subsubsection{RocketChat MCP Server (Enterprise Communication)}

{\small
\begin{longtable}{p{4.5cm} p{10cm}}
\caption{Complete RocketChat MCP Server Tool Catalog} \label{tab:rocketchat_complete} \\
\toprule
\textbf{Tool Name} & \textbf{Description \& Key Parameters} \\
\midrule
\endfirsthead

\multicolumn{2}{c}{\tablename\ \thetable\ -- \textit{Continued from previous page}} \\
\toprule
\textbf{Tool Name} & \textbf{Description \& Key Parameters} \\
\midrule
\endhead

\midrule
\multicolumn{2}{r}{\textit{Continued on next page}} \\
\endfoot

\bottomrule
\endlastfoot

\texttt{send\_user\_message} & Send direct message to user (e.g., 'john.doe', '@john.doe'). \textit{Params:} \texttt{channel} (string, required), \texttt{message} (string, required) \\

\texttt{send\_channel\_message} & Send message to channel (e.g., 'general', '\#general'). \textit{Params:} \texttt{channel} (string, required), \texttt{message} (string, required) \\

\texttt{get\_channels} & List all available channels. \textit{Params:} None \\

\texttt{get\_channel\_messages} & Get recent messages from a channel. \textit{Params:} \texttt{channel} (string, required), \texttt{count} (integer, optional) \\

\texttt{create\_channel} & Create new public or private channel. \textit{Params:} \texttt{name} (string, required), \texttt{description} (string, optional), \texttt{private} (boolean, optional) \\

\texttt{get\_user\_info} & Get information about current authenticated user. \textit{Params:} None \\

\texttt{get\_server\_info} & Get Rocket.Chat server information and statistics. \textit{Params:} None \\

\texttt{search\_messages} & Search for messages across channels. \textit{Params:} \texttt{query} (string, required), \texttt{room\_id} (string, optional) \\

\texttt{get\_direct\_messages} & Get list of direct message conversations. \textit{Params:} None \\

\texttt{get\_users} & Get list of all workspace users. \textit{Params:} None \\

\texttt{join\_channel} & Join a public channel. \textit{Params:} \texttt{channel} (string, required) \\

\texttt{leave\_channel} & Leave a channel. \textit{Params:} \texttt{channel} (string, required) \\

\end{longtable}
}

\subsubsection{Mail System MCP Server}

{\small
\begin{longtable}{p{4.5cm} p{10cm}}
\caption{Complete Mail System MCP Server Tool Catalog} \label{tab:mail_complete} \\
\toprule
\textbf{Tool Name} & \textbf{Description \& Key Parameters} \\
\midrule
\endfirsthead
\multicolumn{2}{c}{\tablename\ \thetable\ -- \textit{Continued from previous page}} \\
\toprule
\textbf{Tool Name} & \textbf{Description \& Key Parameters} \\
\midrule
\endhead
\midrule
\multicolumn{2}{r}{\textit{Continued on next page}} \\
\endfoot
\bottomrule
\endlastfoot

\texttt{send\_email} & Send email with attachments (supports plain text, HTML, and multipart formats). \textit{Params:} \texttt{to} (string, required), \texttt{subject} (string, required), \texttt{body} (string, required), \texttt{attachments} (array, optional) \\

\texttt{draft\_email} & Create draft email without sending. \textit{Params:} \texttt{to} (string, required), \texttt{subject} (string, required), \texttt{body} (string, required) \\

\texttt{read\_email} & Retrieve complete email content by ID with attachment info. \textit{Params:} \texttt{email\_id} (string, required) \\

\texttt{search\_emails} & Search emails by subject, sender, or date range. \textit{Params:} \texttt{query} (string, required), \texttt{from\_date} (string, optional), \texttt{to\_date} (string, optional) \\

\texttt{list\_folders} & List all mail folders (inbox, sent, drafts, etc.). \textit{Params:} None \\

\texttt{move\_email} & Move email to specified folder. \textit{Params:} \texttt{email\_id} (string, required), \texttt{folder} (string, required) \\

\texttt{delete\_email} & Delete email by ID. \textit{Params:} \texttt{email\_id} (string, required) \\

\texttt{mark\_as\_read} & Mark email as read. \textit{Params:} \texttt{email\_id} (string, required) \\

\texttt{mark\_as\_unread} & Mark email as unread. \textit{Params:} \texttt{email\_id} (string, required) \\

\end{longtable}
}

\subsubsection{GitLab MCP Server (Version Control \& CI/CD)}

{\small
\begin{longtable}{p{4.5cm} p{10cm}}
\caption{GitLab MCP Server Tool Catalog (Selected Core Tools)} \label{tab:gitlab_complete} \\
\toprule
\textbf{Tool Name} & \textbf{Description \& Key Parameters} \\
\midrule
\endfirsthead
\multicolumn{2}{c}{\tablename\ \thetable\ -- \textit{Continued from previous page}} \\
\toprule
\textbf{Tool Name} & \textbf{Description \& Key Parameters} \\
\midrule
\endhead
\midrule
\multicolumn{2}{r}{\textit{Continued on next page}} \\
\endfoot
\bottomrule
\multicolumn{2}{l}{\textit{Note: GitLab MCP exposes 65+ tools. See full documentation for complete catalog.}} \\
\endlastfoot

\multicolumn{2}{l}{\textbf{Repository Management}} \\
\texttt{create\_repository} & Create new GitLab project. \textit{Params:} \texttt{name} (string, required), \texttt{description} (string, optional), \texttt{visibility} (string, optional), \texttt{initialize\_with\_readme} (boolean, optional) \\

\texttt{search\_repositories} & Search for GitLab projects. \textit{Params:} \texttt{search} (string, required), \texttt{page} (number, optional), \texttt{per\_page} (number, optional, max 100) \\

\texttt{fork\_repository} & Fork project to your account or namespace. \textit{Params:} \texttt{project\_id} (string, optional), \texttt{namespace} (string, optional) \\

\texttt{get\_repository\_tree} & Get repository file tree structure. \textit{Params:} \texttt{project\_id} (string, optional), \texttt{path} (string, optional), \texttt{ref} (string, optional) \\

\multicolumn{2}{l}{\textbf{File Operations}} \\
\texttt{create\_or\_update\_file} & Create or update single file in project. \textit{Params:} \texttt{project\_id} (string, optional), \texttt{file\_path} (string, required), \texttt{content} (string, required), \texttt{commit\_message} (string, required), \texttt{branch} (string, required) \\

\texttt{get\_file\_contents} & Get contents of file or directory. \textit{Params:} \texttt{project\_id} (string, optional), \texttt{file\_path} (string, required), \texttt{ref} (string, optional) \\

\texttt{push\_files} & Push multiple files in single commit. \textit{Params:} \texttt{project\_id} (string, optional), \texttt{branch} (string, required), \texttt{files} (array, required), \texttt{commit\_message} (string, required) \\

\multicolumn{2}{l}{\textbf{Branch Management}} \\
\texttt{create\_branch} & Create new branch. \textit{Params:} \texttt{project\_id} (string, optional), \texttt{branch} (string, required), \texttt{ref} (string, optional) \\

\texttt{get\_branch\_diffs} & Get changes/diffs between two branches or commits. \textit{Params:} \texttt{project\_id} (string, optional), \texttt{from} (string, required), \texttt{to} (string, required), \texttt{straight} (boolean, optional) \\

\multicolumn{2}{l}{\textbf{Issue Management}} \\
\texttt{create\_issue} & Create new issue in project. \textit{Params:} \texttt{project\_id} (string, optional), \texttt{title} (string, required), \texttt{description} (string, optional), \texttt{assignee\_ids} (array, optional), \texttt{labels} (array, optional) \\

\texttt{list\_issues} & List issues with filters. \textit{Params:} \texttt{project\_id} (string, optional), \texttt{assignee\_id} (string, optional), \texttt{author\_id} (string, optional), \texttt{labels} (array, optional), \texttt{scope} (string, optional) \\

\texttt{update\_issue} & Update existing issue. \textit{Params:} \texttt{project\_id} (string, optional), \texttt{issue\_iid} (string, required), \texttt{title} (string, optional), \texttt{description} (string, optional), \texttt{state\_event} (string, optional) \\

\multicolumn{2}{l}{\textbf{Merge Request Operations}} \\
\texttt{create\_merge\_request} & Create new merge request. \textit{Params:} \texttt{project\_id} (string, optional), \texttt{title} (string, required), \texttt{source\_branch} (string, required), \texttt{target\_branch} (string, required), \texttt{description} (string, optional), \texttt{draft} (boolean, optional) \\

\texttt{merge\_merge\_request} & Merge a merge request. \textit{Params:} \texttt{project\_id} (string, optional), \texttt{merge\_request\_iid} (string, required), \texttt{squash} (boolean, optional), \texttt{should\_remove\_source\_branch} (boolean, optional) \\

\texttt{get\_merge\_request} & Get merge request details. \textit{Params:} \texttt{project\_id} (string, optional), \texttt{merge\_request\_iid} (string, optional), \texttt{source\_branch} (string, optional) \\

\texttt{get\_merge\_request\_diffs} & Get MR changes/diffs. \textit{Params:} \texttt{project\_id} (string, optional), \texttt{merge\_request\_iid} (string, optional), \texttt{source\_branch} (string, optional) \\

\multicolumn{2}{l}{\textbf{Code Review \& Discussion}} \\
\texttt{create\_merge\_request\_thread} & Create new discussion thread on MR. \textit{Params:} \texttt{project\_id} (string, optional), \texttt{merge\_request\_iid} (string, required), \texttt{body} (string, required), \texttt{position} (object, optional) \\

\texttt{create\_draft\_note} & Create draft note for MR. \textit{Params:} \texttt{project\_id} (string, optional), \texttt{merge\_request\_iid} (string, required), \texttt{body} (string, required) \\

\texttt{publish\_draft\_note} & Publish single draft note. \textit{Params:} \texttt{project\_id} (string, optional), \texttt{merge\_request\_iid} (string, required), \texttt{draft\_note\_id} (string, required) \\

\multicolumn{2}{l}{\textbf{Advanced Operations}} \\
\texttt{execute\_graphql} & Execute GitLab GraphQL query. \textit{Params:} \texttt{query} (string, required), \texttt{variables} (object, optional) \\

\texttt{get\_commit\_diff} & Get commit diff details. \textit{Params:} \texttt{project\_id} (string, required), \texttt{commit\_sha} (string, required) \\

\texttt{list\_events} & List project or user events. \textit{Params:} \texttt{target\_type} (string, required), \texttt{target\_id} (string, optional) \\

\end{longtable}
}

\subsubsection{Operations \& IT Management Servers}

{\small
\begin{longtable}{p{4.5cm} p{10cm}}
\caption{Zammad \& Plane MCP Server Tool Catalogs} \label{tab:ops_complete} \\
\toprule
\textbf{Tool Name} & \textbf{Description \& Key Parameters} \\
\midrule
\endfirsthead
\multicolumn{2}{c}{\tablename\ \thetable\ -- \textit{Continued from previous page}} \\
\toprule
\textbf{Tool Name} & \textbf{Description \& Key Parameters} \\
\midrule
\endhead
\midrule
\multicolumn{2}{r}{\textit{Continued on next page}} \\
\endfoot
\bottomrule
\endlastfoot

\multicolumn{2}{l}{\textbf{Zammad MCP (IT Ticketing)}} \\
\texttt{create\_ticket} & Create new support ticket. \textit{Params:} \texttt{title} (string, required), \texttt{description} (string, required), \texttt{priority} (string, optional), \texttt{category} (string, optional) \\

\texttt{get\_ticket} & Retrieve ticket details by ID. \textit{Params:} \texttt{ticket\_id} (string, required) \\

\texttt{update\_ticket} & Update ticket status, priority, or assignee. \textit{Params:} \texttt{ticket\_id} (string, required), \texttt{status} (string, optional), \texttt{priority} (string, optional), \texttt{assignee\_id} (string, optional) \\

\texttt{search\_tickets} & Search tickets by criteria. \textit{Params:} \texttt{query} (string, required), \texttt{status} (string, optional), \texttt{assignee} (string, optional) \\

\texttt{add\_note} & Add internal or public note to ticket. \textit{Params:} \texttt{ticket\_id} (string, required), \texttt{note} (string, required), \texttt{internal} (boolean, optional) \\

\texttt{assign\_agent} & Assign ticket to support agent. \textit{Params:} \texttt{ticket\_id} (string, required), \texttt{agent\_id} (string, required) \\

\texttt{close\_ticket} & Close resolved ticket. \textit{Params:} \texttt{ticket\_id} (string, required), \texttt{resolution} (string, optional) \\

\texttt{get\_ticket\_history} & Retrieve complete ticket activity log. \textit{Params:} \texttt{ticket\_id} (string, required) \\

\texttt{list\_agents} & Get available support agents. \textit{Params:} None \\

\texttt{get\_priorities} & List ticket priority levels. \textit{Params:} None \\

\texttt{get\_categories} & List ticket categories. \textit{Params:} None \\

\multicolumn{2}{l}{\textbf{Plane MCP (Jira-style Project Management)}} \\
\texttt{get\_projects} & List all workspace projects. \textit{Params:} \texttt{workspace\_slug} (string, optional) \\

\texttt{create\_project} & Create new project. \textit{Params:} \texttt{workspace\_slug} (string, required), \texttt{name} (string, required), \texttt{description} (string, optional) \\

\texttt{list\_issues} & List project issues. \textit{Params:} \texttt{workspace\_slug} (string, required), \texttt{project\_id} (string, required), \texttt{state} (string, optional), \texttt{assignee} (string, optional) \\

\texttt{create\_issue} & Create new issue. \textit{Params:} \texttt{workspace\_slug} (string, required), \texttt{project\_id} (string, required), \texttt{name} (string, required), \texttt{description} (string, optional) \\

\texttt{update\_issue} & Update issue details. \textit{Params:} \texttt{workspace\_slug} (string, required), \texttt{project\_id} (string, required), \texttt{issue\_id} (string, required), updates (object, required) \\

\texttt{create\_state} & Create workflow state. \textit{Params:} \texttt{workspace\_slug} (string, required), \texttt{project\_id} (string, required), \texttt{name} (string, required), \texttt{color} (string, optional) \\

\texttt{create\_label} & Create issue label. \textit{Params:} \texttt{workspace\_slug} (string, required), \texttt{project\_id} (string, required), \texttt{name} (string, required), \texttt{color} (string, optional) \\

\texttt{create\_cycle} & Create sprint cycle. \textit{Params:} \texttt{workspace\_slug} (string, required), \texttt{project\_id} (string, required), \texttt{name} (string, required), \texttt{start\_date} (string, required), \texttt{end\_date} (string, required) \\

\texttt{create\_module} & Create project module. \textit{Params:} \texttt{workspace\_slug} (string, required), \texttt{project\_id} (string, required), \texttt{name} (string, required) \\

\end{longtable}
}

\subsubsection{Human Resources, Storage, \& Business Servers}

{\small
\begin{longtable}{p{4.5cm} p{10cm}}
\caption{Frappe HR, OwnCloud, Mongoose, \& Dolibarr MCP Server Catalogs} \label{tab:business_complete} \\
\toprule
\textbf{Tool Name} & \textbf{Description \& Key Parameters} \\
\midrule
\endfirsthead
\multicolumn{2}{c}{\tablename\ \thetable\ -- \textit{Continued from previous page}} \\
\toprule
\textbf{Tool Name} & \textbf{Description \& Key Parameters} \\
\midrule
\endhead
\midrule
\multicolumn{2}{r}{\textit{Continued on next page}} \\
\endfoot
\bottomrule
\endlastfoot

\multicolumn{2}{l}{\textbf{Frappe HR MCP}} \\
\texttt{get\_employees} & Retrieve employee list with filters. \textit{Params:} \texttt{department} (string, optional), \texttt{designation} (string, optional), \texttt{status} (string, optional) \\

\texttt{get\_employee} & Get employee details by ID. \textit{Params:} \texttt{employee\_id} (string, required) \\

\texttt{create\_employee} & Onboard new employee. \textit{Params:} \texttt{first\_name} (string, required), \texttt{last\_name} (string, required), \texttt{email} (string, required), \texttt{department} (string, required) \\

\texttt{get\_departments} & List organization departments. \textit{Params:} None \\

\texttt{get\_attendance} & Retrieve attendance records. \textit{Params:} \texttt{employee\_id} (string, optional), \texttt{from\_date} (string, optional), \texttt{to\_date} (string, optional) \\

\texttt{create\_leave\_request} & Submit leave application. \textit{Params:} \texttt{employee\_id} (string, required), \texttt{leave\_type} (string, required), \texttt{from\_date} (string, required), \texttt{to\_date} (string, required) \\

\texttt{get\_salary\_slips} & Retrieve payroll information. \textit{Params:} \texttt{employee\_id} (string, optional), \texttt{month} (string, optional) \\

\texttt{get\_timesheets} & Get time tracking data. \textit{Params:} \texttt{employee\_id} (string, optional), \texttt{project} (string, optional) \\

\texttt{get\_hrms\_report} & Generate HR analytics reports. \textit{Params:} \texttt{report\_type} (string, required), \texttt{filters} (object, optional) \\

\multicolumn{2}{l}{\textbf{OwnCloud MCP (Document Management)}} \\
\texttt{list\_files} & List files and folders in path. \textit{Params:} \texttt{path} (string, optional) \\

\texttt{upload\_file} & Upload file to OwnCloud. \textit{Params:} \texttt{local\_path} (string, required), \texttt{remote\_path} (string, required) \\

\texttt{download\_file} & Download file from OwnCloud. \textit{Params:} \texttt{remote\_path} (string, required), \texttt{local\_path} (string, required) \\

\texttt{read\_file\_content} & Read and return text file content. \textit{Params:} \texttt{remote\_path} (string, required) \\

\texttt{search\_files} & Search files by name. \textit{Params:} \texttt{query} (string, required) \\

\texttt{create\_folder} & Create new folder. \textit{Params:} \texttt{path} (string, required) \\

\texttt{delete\_file} & Delete file or folder. \textit{Params:} \texttt{path} (string, required) \\

\texttt{get\_storage\_info} & Get quota and usage statistics. \textit{Params:} None \\

\multicolumn{2}{l}{\textbf{Mongoose MCP (Employee Database)}} \\
\texttt{create\_record} & Insert new employee database record. \textit{Params:} \texttt{collection} (string, required), \texttt{data} (object, required) \\

\texttt{query\_database} & Query database with filters. \textit{Params:} \texttt{collection} (string, required), \texttt{filter} (object, required), \texttt{projection} (object, optional) \\

\texttt{update\_employee} & Update employee information. \textit{Params:} \texttt{employee\_id} (string, required), \texttt{updates} (object, required) \\

\texttt{delete\_record} & Remove database record. \textit{Params:} \texttt{collection} (string, required), \texttt{id} (string, required) \\

\texttt{aggregate\_data} & Perform aggregation queries. \textit{Params:} \texttt{collection} (string, required), \texttt{pipeline} (array, required) \\

\multicolumn{2}{l}{\textbf{Dolibarr CRM MCP}} \\
\texttt{get\_customers} & List all CRM customers. \textit{Params:} \texttt{status} (string, optional), \texttt{search} (string, optional) \\

\texttt{create\_customer} & Create new customer record. \textit{Params:} \texttt{name} (string, required), \texttt{email} (string, optional), \texttt{phone} (string, optional) \\

\texttt{get\_products} & List product catalog. \textit{Params:} \texttt{category} (string, optional) \\

\texttt{create\_product} & Add new product. \textit{Params:} \texttt{name} (string, required), \texttt{price} (number, required), \texttt{description} (string, optional) \\

\texttt{create\_order} & Create sales order. \textit{Params:} \texttt{customer\_id} (string, required), \texttt{products} (array, required) \\

\texttt{create\_invoice} & Generate customer invoice. \textit{Params:} \texttt{order\_id} (string, required), \texttt{due\_date} (string, optional) \\

\texttt{create\_holiday\_request} & Submit HR leave request. \textit{Params:} \texttt{user\_id} (string, required), \texttt{start\_date} (string, required), \texttt{end\_date} (string, required) \\

\end{longtable}
}

\subsubsection{Utility Servers}

{\small
\begin{longtable}{p{4.5cm} p{10cm}}
\caption{Utility MCP Server Tool Catalog (File System, Browser, Bash)} \label{tab:utility_complete} \\
\toprule
\textbf{Tool Name} & \textbf{Description \& Key Parameters} \\
\midrule
\endfirsthead
\multicolumn{2}{c}{\tablename\ \thetable\ -- \textit{Continued from previous page}} \\
\toprule
\textbf{Tool Name} & \textbf{Description \& Key Parameters} \\
\midrule
\endhead
\bottomrule
\endlastfoot

\multicolumn{2}{l}{\textbf{File System MCP}} \\
\texttt{read\_file} & Read file contents. \textit{Params:} \texttt{path} (string, required) \\

\texttt{write\_file} & Write data to file. \textit{Params:} \texttt{path} (string, required), \texttt{content} (string, required) \\

\texttt{list\_directory} & List directory contents. \textit{Params:} \texttt{path} (string, required) \\

\texttt{create\_directory} & Create new directory. \textit{Params:} \texttt{path} (string, required) \\

\multicolumn{2}{l}{\textbf{Web Browser MCP}} \\
\texttt{browse\_url} & Navigate to URL and retrieve content. \textit{Params:} \texttt{url} (string, required) \\

\texttt{search\_web} & Perform web search. \textit{Params:} \texttt{query} (string, required), \texttt{num\_results} (integer, optional) \\

\texttt{extract\_links} & Extract hyperlinks from page. \textit{Params:} \texttt{url} (string, required), \texttt{filter} (string, optional) \\

\multicolumn{2}{l}{\textbf{Bash Shell MCP}} \\
\texttt{execute\_command} & Execute shell command. \textit{Params:} \texttt{command} (string, required), \texttt{timeout} (integer, optional) \\

\texttt{run\_script} & Run bash script. \textit{Params:} \texttt{script\_path} (string, required), \texttt{args} (array, optional) \\

\texttt{get\_environment} & Get environment variables. \textit{Params:} \texttt{var\_name} (string, optional) \\

\end{longtable}
}

\subsection{Limitations}
\label{app:lim}

While EnterpriseLab demonstrates strong performance on enterprise agentic tasks, we acknowledge the following scope and directions for future work: (1) Our platform focuses on tool-based agentic environments with API interactions; extending to UI-based environments with visual grounding represents a natural next step. (2) We achieve competitive performance on enterprise benchmarks, matching or exceeding proprietary models on several tasks, though state-of-the-art models like Gemini-2.5-Pro maintain advantages on certain complex scenarios. (3) Agentic GRPO performs optimally with reasonably capable base models; for weaker initializations, additional supervised fine-tuning provides better learning dynamics. (4) Task generation quality scales with environment complexity—richer tool dependencies and workflow diversity enable more comprehensive training data synthesis. (5) Our evaluation focuses on tool-using agents in enterprise settings; validating Agentic GRPO on other domains such as mathematical reasoning or code generation remains valuable future work.

\subsection{Prompts}
\label{app:prompt}

\subsubsection{Prompt for Trajectory Level Thought Generation}
\label{app:thought_prompt}
\begin{tcolorbox}[
    enhanced,
    breakable,
    width=\linewidth,
    title=Prompt for Trajectory Level Thought Generation,
    fonttitle=\bfseries\normalsize,  
    colframe=blue!75!black,
    colback=white!10!white,
    coltitle=white,
    colbacktitle=blue!85!black,
    boxrule=0.2mm,
    sharp corners,
    shadow={1mm}{-1mm}{0mm}{black!50!white},
    attach boxed title to top left={yshift=-2mm, xshift=3mm},  
    boxed title style={sharp corners, size=small},
    top=3mm,  
    bottom=2mm  
]

\small
\ttfamily 
\setlength{\parindent}{0pt} 
\setlength{\parskip}{0.5em} 

You are an AI agent. Explain your NEXT action in natural language.

\textbf{CONTEXT:}
{context}

\textbf{CURRENT ACTION:}
Operation: {step.tool\_name.replace('\_', ' ').title()}
With Data: {key\_values\_str}

\textbf{Full Inputs:}
{input\_preview}

\textbf{NEXT ACTION:}
Operation: {step.tool\_name.replace('\_', ' ').title()}
With Data: {key\_values\_str}

\textbf{Full Inputs:}
{input\_preview}

RULES:
1. Write ONE sentence (first-person: "I am...", "I will...")

2. Use BUSINESS LANGUAGE - no tool names, no technical jargon

    - BAD: "I will execute owncloud\_create\_folder"

    - GOOD: "I am creating a new folder in cloud storage"

3. Reference SPECIFIC data values (IDs, names, paths)

4. Explain WHAT and WHY in plain terms

YOUR RESPONSE:
\end{tcolorbox}
\subsubsection{Prompt for High Level Task Generation}
\label{app:high_prompt}

\begin{tcolorbox}[
    enhanced,
    breakable,
    width=\linewidth,
    title=Prompt for High Level Task Generation,
    fonttitle=\bfseries\normalsize,  
    colframe=blue!75!black,
    colback=white!10!white,
    coltitle=white,
    colbacktitle=blue!85!black,
    boxrule=0.2mm,
    sharp corners,
    shadow={1mm}{-1mm}{0mm}{black!50!white},
    attach boxed title to top left={yshift=-2mm, xshift=3mm},  
    boxed title style={sharp corners, size=small},
    top=3mm,  
    bottom=2mm  
]

\small
\ttfamily 
\setlength{\parindent}{0pt} 
\setlength{\parskip}{0.5em} 

Create a natural user instruction from this execution trace.

\textbf{DOMAIN:} {domain}

\textbf{TRACE (what actually happened):}
\{trace\_text\}

\textbf{REQUIRED DATA (must appear in instruction):}
\{data\_block\}

RULES:
1. Write as a realistic business request (2-3 sentences)

2. Include ALL specific data values (IDs, names, paths, etc.)

3. Use business language, not technical terms

4. Do NOT say "use tool X" or "call API Y"

5. Be GROUNDED - if tools created data, ask to create; if retrieved, ask to retrieve

OUTPUT (JSON):
{{
  "instruction": "Full natural language request with all data",
  "success\_criteria": [
    "Criterion 1",
    "Criterion 2"
  ]
}}

YOUR RESPONSE: 
\end{tcolorbox}
\subsubsection{Prompt for Agentic GRPO Training rollout generation}
\label{app:agrpo_prompt}

During the Agentic GRPO training phase (Section~\ref{sec:scapia-training}), the agent generates trajectories by interacting with the environment using the ReAct prompting format. The following system prompt is provided to the policy model during rollouts to enforce structured reasoning and action execution. This prompt explicitly defines the required format (Thought, Action, Action Input, Observation, Final Answer) and provides concrete examples to guide the model in generating valid tool-calling trajectories that can be parsed and executed by the environment.

\begin{tcolorbox}[
    enhanced,
    breakable,
    width=\linewidth,
    title=System Prompt for Agentic GRPO Training,
    fonttitle=\bfseries\normalsize,  
    colframe=blue!75!black,
    colback=white!10!white,
    coltitle=white,
    colbacktitle=blue!85!black,
    boxrule=0.2mm,
    sharp corners,
    shadow={1mm}{-1mm}{0mm}{black!50!white},
    attach boxed title to top left={yshift=-2mm, xshift=3mm},  
    boxed title style={sharp corners, size=small},
    top=3mm,  
    bottom=2mm  
]

\small
\ttfamily 
\setlength{\parindent}{0pt} 
\setlength{\parskip}{0.5em} 

You are an AI assistant that helps users by using tools. You MUST follow the ReAct format exactly.

\textbf{AVAILABLE TOOLS:}\{\}

\textbf{STRICT FORMAT REQUIREMENTS:}

You MUST use this EXACT format for every response:

Thought: [Your reasoning about what to do next]\\
Action: [The tool name to use]\\
Action Input: \{"param1": "value1", "param2": "value2"\}\\
\\
After the tool executes, you will see:\\
\\
Observation: [Tool output]

Then continue with another Thought/Action cycle, OR provide final answer:

Final Answer: [Your final response to the user]

\textbf{CRITICAL RULES:}
1. ALWAYS start with "Thought:" (capital T, colon, space)\\
2. ALWAYS try to find the action inputs from the observation or previous steps or user query\\
3. Action name must EXACTLY match a tool name from the list above\\
4. Action Input MUST be valid JSON with double quotes\\
5. Do NOT use XML tags like <think>, <action>, etc.\\
6. Do NOT add extra formatting or markdown\\
7. Each line starts with the keyword (Thought:, Action:, etc.)

\textbf{EXAMPLE 1 - Reading an email:}

User: Read email with ID email\_123

Thought: I need to retrieve the email content using the read\_email tool.\\
Action: read\_email\\
Action Input: \{"email\_id": "email\_123"\}\\
Observation: \{"subject": "Meeting Update", "body": "The meeting is at 3pm", "sender": "alice@company.com"\}

Thought: I have retrieved the email content successfully.\\
Final Answer: The email with ID email\_123 has subject "Meeting Update" from [alice@company.com](mailto:alice@company.com). The body says: "The meeting is at 3pm"

\textbf{EXAMPLE 2 - Multi-step GitHub task:}

User: List all issues in the repository called "myproject"

Thought: I need to get the list of issues for the repository named "myproject".\\
Action: github\_list\_issues\_of\_repository\\
Action Input: \{"repo\_name": "myproject"\}\\
Observation: [\{"id": "issue\_1", "title": "Bug in login", "status": "Open"\}, \{"id": "issue\_2", "title": "Add feature", "status": "Closed"\}]

Thought: I have retrieved the list of issues. There are 2 issues total.\\
Final Answer: The repository "myproject" has 2 issues:\\
1. Issue issue\_1: "Bug in login" (Open)\\
2. Issue issue\_2: "Add feature" (Closed)

\textbf{EXAMPLE 3 - Creating and Updating:}

User: Create a product called "Widget Pro" with ID prod\_123, then update its price to \$29.99

Thought: The user wants to create and then update a product. I should first create the product 'prod\_123'. I will verify the creation was successful before updating the price.\\
Action: create\_product\\
Action Input: \{"product\_id": "prod\_123", "product\_name": "Widget Pro"\}\\
Observation: Product created successfully with ID prod\_123

Thought: The product 'prod\_123' was created successfully. Now I can proceed to update its price to 29.99.\\
Action: update\_product\\
Action Input: \{"product\_id": "prod\_123", "actual\_price": "29.99"\}

Thought: Both steps completed successfully.\\
Final Answer: I created the product "Widget Pro" with ID prod\_123 and updated its price to \$29.99.

\textbf{EXAMPLE 4 - Handling Errors:}

User: Get details for ticket T-999

Thought: I need to fetch the ticket details for T-999.\\
Action: get\_it\_ticket\\
Action Input: \{"id": "T-999"\}\\
Observation: Error: Ticket T-999 not found.

Thought: The ticket T-999 was not found. I might have the wrong ID, or I need to search for tickets assigned to me to find the correct ID. I will list my tickets first.\\
Action: list\_it\_tickets\_assigned\_to\_me\\
Action Input: \{"emp\_id": "current\_user"\}

Now, follow this format exactly for all tasks!
\end{tcolorbox}


\end{document}